\documentclass{article}

\PassOptionsToPackage{numbers, compress}{natbib}

\usepackage[final]{neurips_2025}

\usepackage[utf8]{inputenc} 
\usepackage[T1]{fontenc}    
\usepackage{hyperref}       
\usepackage{url}            
\usepackage{booktabs}       
\usepackage{amsfonts}       
\usepackage{nicefrac}       
\usepackage{microtype}      
\usepackage{xcolor}         

\usepackage{graphicx}
\usepackage{multirow}
\usepackage{makecell} 
\usepackage{kotex}
\usepackage{amsmath}

\usepackage{pifont}

\usepackage{caption}
\usepackage{wrapfig}
\usepackage{adjustbox}
\usepackage{subcaption}
\usepackage[table]{xcolor}
\usepackage{array}

\definecolor{hi}{RGB}{180,210,255}   
\definecolor{mid}{RGB}{255,240,190}  
\definecolor{low}{RGB}{255,210,210}  

\newcommand{\scorecell}[1]{%
  \begingroup
    \def\val{#1}%
    \ifdim \val pt > 0.95pt
      \cellcolor{hi}{#1}
    \else\ifdim \val pt > 0.85pt
      \cellcolor{mid}{#1}
    \else
      \cellcolor{low}{#1}
    \fi\fi
  \endgroup
}

\newcommand{\retr}[2]{%
  & #1 & \scorecell{#2} \\
}

\title{
RadZero: Similarity-Based Cross-Attention for Explainable Vision-Language Alignment in \\Chest X-ray with Zero-Shot Multi-Task Capability
}

\author{%
  Jonggwon Park, Byungmu Yoon, Soobum Kim, Kyoyun Choi\thanks{Corresponding author} \\
  DEEPNOID Inc.\\
  Seoul, South Korea \\
  \texttt{\{jgpark, bmyoon, soobumk, kychoi\}@deepnoid.com} \\
}

\begin{document}
\maketitle

\begin{abstract}
Recent advancements in multimodal models have significantly improved vision-language (VL) alignment in radiology. 
However, existing approaches struggle to effectively utilize complex radiology reports for learning and offer limited interpretability through attention probability visualizations.
To address these challenges, we introduce \textbf{RadZero}, a novel framework for VL alignment in chest X-ray with zero-shot multi-task capability.
A key component of our approach is \textbf{VL-CABS} (\textbf{V}ision-\textbf{L}anguage \textbf{C}ross-\textbf{A}ttention \textbf{B}ased on \textbf{S}imilarity), which aligns text embeddings with local image features for interpretable, fine-grained VL reasoning.
RadZero leverages large language models to extract concise semantic sentences from radiology reports and employs multi-positive contrastive training to effectively capture relationships between images and multiple relevant textual descriptions. 
It uses a pre-trained vision encoder with additional trainable Transformer layers, allowing efficient high-resolution image processing. 
By computing similarity between text embeddings and local image patch features, VL-CABS enables zero-shot inference with similarity 
probability for classification, and pixel-level VL similarity maps for grounding and segmentation. 
Experimental results on public chest radiograph benchmarks show that RadZero outperforms state-of-the-art methods in zero-shot classification, grounding, and segmentation. 
Furthermore, VL similarity map analysis highlights the potential of VL-CABS for improving explainability in VL alignment. 
Additionally, qualitative evaluation demonstrates RadZero's capability for open-vocabulary semantic segmentation, further validating its effectiveness in medical imaging.
Code is available at \href{https://github.com/deepnoid-ai/RadZero}{https://github.com/deepnoid-ai/RadZero}.

\end{abstract}    
\section{Introduction}
\label{sec:intro}

Recent advancements in deep learning have significantly impacted medical imaging, 
leading to 
numerous studies on computer-aided diagnosis \cite{cad_1, cad_2, cad_3, thorax_cls}. 
However, acquiring high-quality manual annotations remains a key challenge. 
In contrast, vision-language (VL) models (VLMs) in the natural image domain \cite{clip, siglip, lit} have reduced reliance on manual labeling by learning from image-text pairs without explicit supervision, achieving strong zero-shot performance in tasks like classification and retrieval.
Building on this progress, VLMs have been increasingly explored in medical imaging, including chest X-rays (CXRs). 
Several studies have demonstrated effective representation learning \cite{medklip, kad, convirt} and zero-shot capabilities \cite{carzero, gzsl} without task-specific annotations.

\begin{figure}[t]
    \begin{center}
    \includegraphics[width=\linewidth]{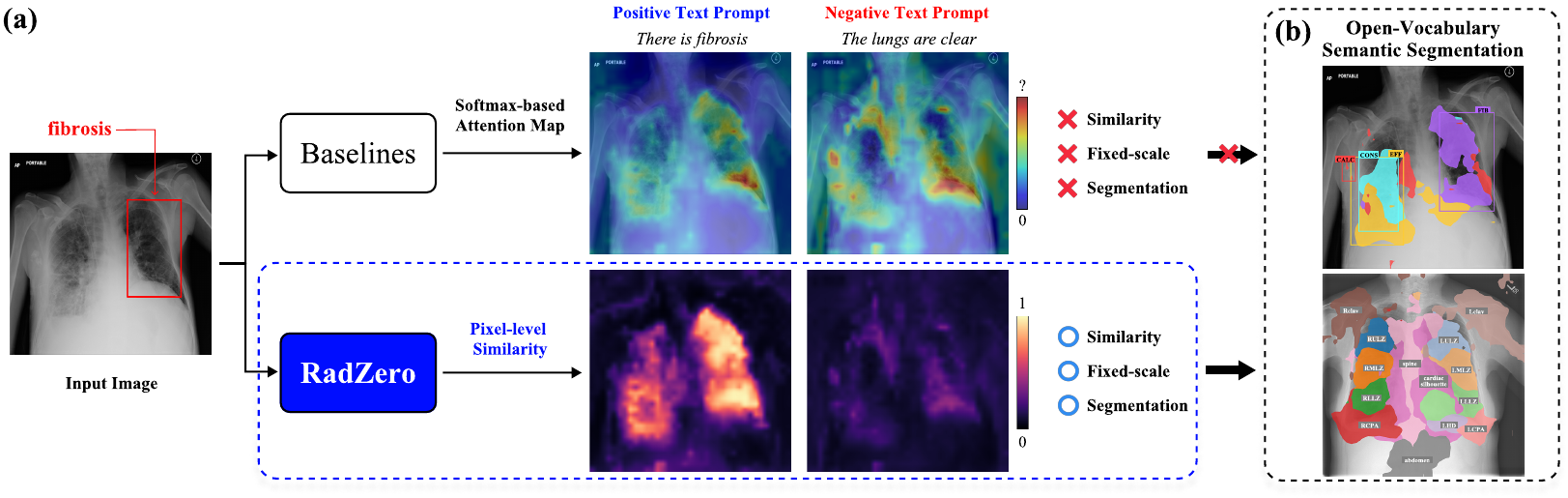}
    \end{center}
    \caption{
Comparison of attention maps and the proposed VL similarity map for visualizing VL alignment.
(a) While traditional attention maps inevitably exhibit high values at certain points due to softmax activation, the proposed VL similarity maps yield low values for unrelated image-text pair.
(b) Their fixed scale, originating from cosine similarity, enables open-vocabulary semantic segmentation through simple thresholding.
}
    \label{fig:introduction}
\end{figure}

Despite these advance, current medical 
VLMs underutilize the rich semantics of radiology image–report pairs.
Prior methods—such as word-level alignment \cite{gloria, biovil-t}, 
clinical entity extraction \cite{kad, medklip}, 
and using large language model (LLM) prompts \cite{carzero}—face limitations, struggling with poor text embedding segmentation and inefficient training due to random sampling. 
An effective solution should 
1) decompose reports into semantically minimal, clinically meaningful sentences, 
2) embed each sentence independently, and 
3) learn from multiple sentence–image pairs per study to fully leverage supervision.

Moreover, reliable explainability is critical for the clinical adoption of medical VLMs.
Attention maps are used as explainable features by most of the recent research \cite{medklip, carzero, kad}.
However, while an attention map (Figure \ref{fig:introduction} (a), top) can indicate where the model is focusing, it does not provide an explanation for why it is attending to those regions.
This limitation can be addressed by computing pixel-level image–text similarity, which enables more fine-grained and transparent explanations.

\begin{wrapfigure}{rt}{0.5\textwidth}
  \centering
  \includegraphics[width=0.48\textwidth]{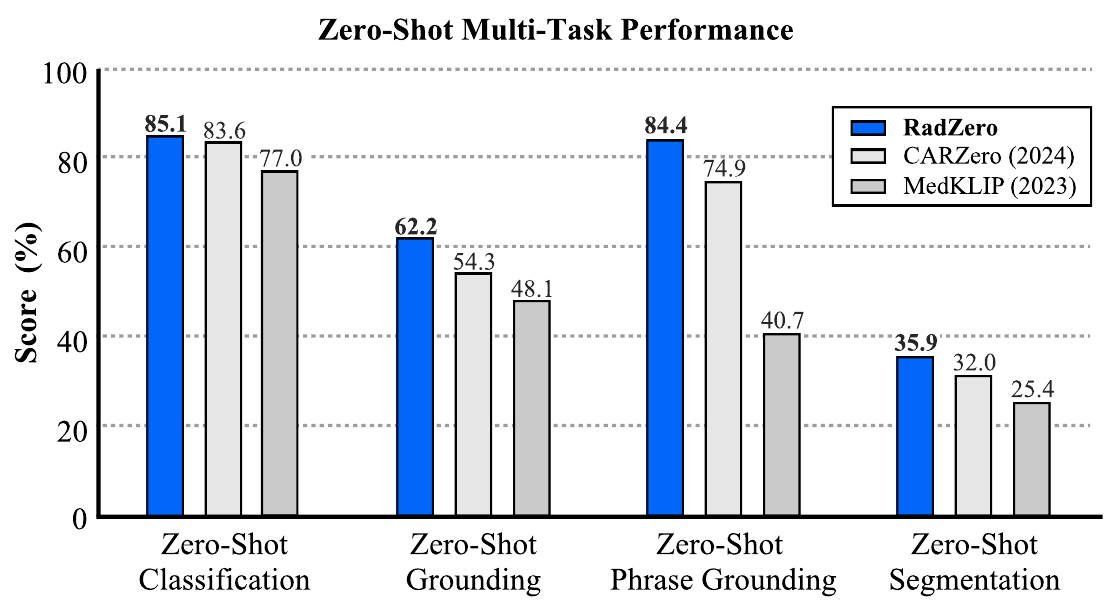}
  \caption{Zero-shot multi-task performance. Each score is averaged over multiple datasets per task.}
  \label{fig:introduction-performance}
\end{wrapfigure}

To overcome these shortcomings, 
we propose \textbf{RadZero}, a novel VL alignment framework for chest X-ray with zero-shot multi-task capabilities.
RadZero employs multi-positive contrastive learning \cite{uniclip} to incorporate multiple sentences per image-report pair. 
RadZero’s core innovation is \textbf{V}ision-\textbf{L}anguage \textbf{C}ross-\textbf{A}ttention \textbf{B}ased on \textbf{S}imilarity (\textbf{VL-CABS}), which directly computes cosine similarity between text descriptions and local image patches. 
Unlike traditional attention maps, the resulting
VL similarity maps offer clearer visual reasoning 
by maintaining low values for unrelated image-text pairs
(Figure \ref{fig:introduction} (a), bottom).
This enhances interpretability and 
enables open-vocabulary semantic segmentation via simple thresholding.
RadZero also supports high-resolution inputs by freezing a pre-trained image encoder \cite{lit} and adding trainable Transformer layers
\cite{dinotxt},
further boosting performance 
on fine-grained zero-shot tasks.
Experiments on public 
chest radiograph 
benchmarks demonstrate that RadZero outperforms state-of-the-art (SOTA) models in 
various zero-shot tasks (Figure~\ref{fig:introduction-performance}),
while qualitative analyses reveal its enhanced explainability and potential for open-vocabulary semantic segmentation.

\section{Related Works}
\label{sec:related}

\subsection{General vision-language alignment}
Contrastive learning for vision–language alignment with large-scale image–text pairs has been actively studied.
CLIP \cite{clip} demonstrated that directly aligning images and text enables strong zero-shot classification.
LiT \cite{lit} proposed freezing the pre-trained vision encoder during contrastive training, preserving fine-grained visual features and further enhancing zero-shot performance.
\texttt{dino.txt} \cite{dinotxt} extended this framework by adding Transformer \cite{transformer} layers on top of a pre-trained DINOv2 \cite{dinov2}, training only a lightweight module while keeping the vision encoder frozen.
Additionally, it fused global and patch-averaged embeddings, enabling patch-level similarity computation with text and supporting open-vocabulary semantic segmentation.
UniCLIP \cite{uniclip} introduced a multi-positive NCE (MP-NCE) loss, which independently evaluates multiple positive pairs per image. 
Building on these advances, our approach integrates a frozen, fine-grained vision encoder with trainable Transformer layers, following LiT and \texttt{dino.txt}.
We also adopt MP-NCE loss to align images with multiple text representations effectively.

\subsection{Vision-language alignment 
in chest X-ray} 
\label{sec:vla_rad}

Since the introduction of contrastive learning in radiology \cite{convirt}, aligning CXR images with radiology reports has become an active research area.
GLoRIA \cite{gloria} focused on local alignment using cross-attention between word-level text embeddings and patch-level image features.
MGCA \cite{mgca} employed both report-level and token-level embeddings to extract multi-granular features, and
BioViL-T \cite{biovil-t}
similarly relied on token-level embeddings.
Nevertheless, segmenting reports into individual words or tokens often fails to capture their full semantic meaning.

Due to the complexity of medical image–report relationships, alignment interpretability is essential for clinical use and is commonly addressed using attention maps.
MedKLIP \cite{medklip} and KAD \cite{kad}, for example, used RadGraph \cite{radgraph} to extract report features and employed attention maps for tasks such as grounding and segmentation.
In addition to VL alignment,
G2D \cite{g2d} aggregated attention maps in addition to VL alignment
to generate pseudo masks, which were used as pixel-wise pretext supervisory signals during pre-training.
CARZero \cite{carzero} also used attention maps when leveraging cross-attention alignment 
for zero-shot tasks, 
incorporating LLM-based prompt alignment to standardize reports.
Despite their utility, attention maps have limitations: they often highlight irrelevant regions due to softmax activation, but removing softmax is not ideal as raw logits are unnormalized and uncentered.
Additionally, variation in the norms of query and key embeddings leads to inconsistent similarity values across different image-text pairs.
An example and a detailed discussion of 
the limited explainability of attention maps
are provided in Appendix \ref{appendix:attenttion_map}.
In contrast, our approach enhances explainability 
with VL-CABS, aligning
visual patches and text embeddings.
The resulting maps offer intuitive and consistent measures of fine-grained image-text similarity.

\section{Methods}
\label{sec:methods}

\subsection{Finding-sentence extraction}
\label{sec:finding_extraction}
\begin{figure*}[t]
    \begin{center}
    \scalebox{1.0}{
        \includegraphics[width=\linewidth]{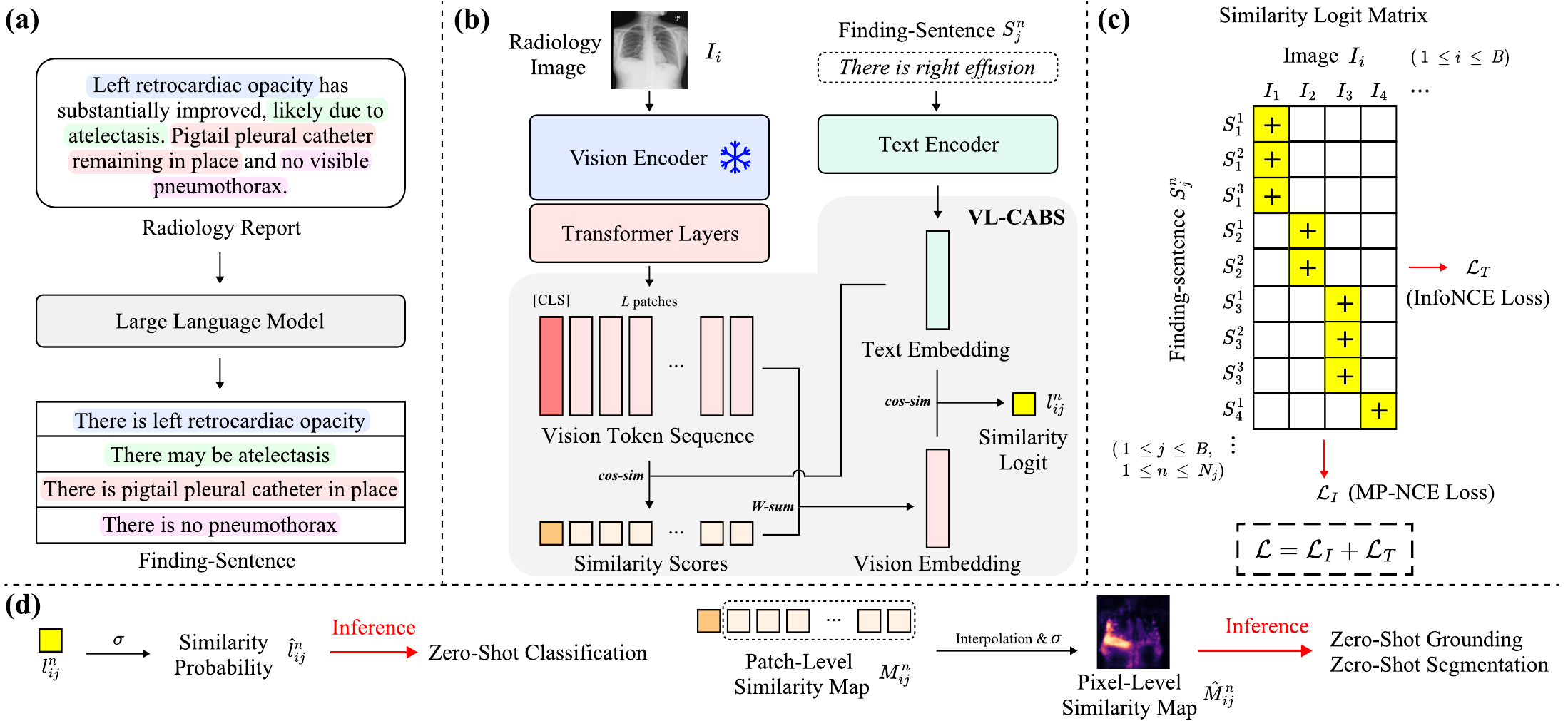}
    }
    \vspace{-0.5cm}
    \end{center}
    \caption{
    The overall framework of RadZero.
    (a) Finding-sentence extraction using an LLM.
    (b) Computation of the similarity logit, \(l_{ij}^n\), between image \(I_{i}\) and finding-sentence \(S_{j}^n\).
    \textbf{\textit{W-sum}} and \textbf{\textit{cos-sim}}
    denote weighted sum and cosine similarity, respectively.
    (c) Computation of MP-NCE loss (\(\mathcal{L}_I\)) and InfoNCE loss (\(\mathcal{L}_T\)) from the similarity logit matrix.
    (d) Zero-shot inference pipeline.
    }
    \vspace{-0.3cm}
    \label{fig:method}
\end{figure*}

Radiology reports contain diverse types of information, including clinical history, observations, comparative analysis with prior studies, and diagnostic impressions. 
Encoding the entire report into a single text embedding often fails to capture this complexity.
CARZero \cite{carzero} addressed this by using an LLM to extract relevant sentences and introducing a prompt alignment strategy based on the template 
“There is [disease]” for consistency between training and inference. 
Similarly, we use an LLM to extract such sentences, which we refer to as 
\textit{finding-sentences}.
These are generated using a prompt that follows a predefined structure, such as ``There is [finding] of [location],'' and are segmented into minimal semantic units containing the finding name, presence (or uncertainty), and location.
The full prompt is provided in Appendix \ref{appendix:prompt}.
Each image is paired with multiple finding-sentences during training, as illustrated in Figure \ref{fig:method} (a). 
For zero-shot inference, we apply prompt alignment by prepending ``There is'' to text descriptions of findings and anatomical regions.

\subsection{Vision-language alignment with similarity based cross-attention}
\subsubsection{Model architecture}
To leverage the advantages of vision encoder pre-training, we adopt the approach of LiT \cite{lit} by freezing a pre-trained vision encoder in contrastive learning. 
In Vision Transformers \cite{vit} such as DINOv2 \cite{dinov2}, interpolating the positional embeddings allows for increased input image resolution \cite{eagle}. 
Building on this property, we train our model with high-resolution images.
To embed the output of the vision encoder, we add $K$ trainable Transformer layers, as proposed by \citet{dinotxt}. 
For the text encoder, we use a pre-trained Sentence-BERT 
\cite{mpnet}, which is fine-tuned during training, 
to extract embeddings for each finding-sentence.
The model architecture is illustrated in Figure \ref{fig:method} (b).

\subsubsection{Vision-language cross-attention based on similarity}
\label{sec:similarity-based-cross-attention}

We propose
VL-CABS
(Vision-Language Cross-Attention Based on Similarity),
a cosine similarity-based cross-attention mechanism
for computing similarity logits.
By directly employing cosine similarity between 
the text and visual patch embeddings, 
we obtain VL similarity scores
that are well-defined in range and centered at zero.
This consistent scaling allows for 
fair comparisons across different image–text pairs 
and significantly enhances explainability 
through the visualization of VL similarity maps.
It also enables single thresholding,
offering new possibilities for open-vocabulary semantic segmentation.

The proposed method operates on a mini-batch of size $B$, pairing each image $I_i$ ($i = 1,\dots,B$) with $N_i$ associated finding-sentences $\{S_i^n\}_{n=1}^{N_i}$. Each image is processed by a vision encoder $f_v$ followed by 
trainable
layers $f_a$, yielding a sequence of embeddings $[\,v_{i0},\,v_{i1},\dots,v_{iL}\,] = f_a(f_v(I_i)),$
where $v_{i0} \in \mathbb{R}^D$ corresponds to the \texttt{[CLS]} token and $\{v_{ik}\}_{k=1}^L\subset \mathbb{R}^D$ are patch embeddings, with $D$ denoting the embedding dimension and $L$ the total number of patches.
Each finding-sentence $S_j^n$ is encoded into a sentence-level
embedding
$t_j^n = f_t(S_j^n) \in \mathbb{R}^D$ using a text encoder $f_t$.

To compute VL similarity, we ℓ2-normalize all embeddings as $\bar{v}_{ik} = v_{ik} / \|v_{ik}\|_2$ and $\bar{t}_j^n = t_j^n / \|t_j^n\|_2$, and calculate the scaled cosine similarity between each patch and sentence as
$
s_{ijk}^n = <\bar{v}_{ik}, \bar{t}_j^n> \cdot \exp(\tau) \quad (k = 0, \dots, L), 
$
where $\tau$ is a learnable temperature. These scores are converted into attention weights using softmax over the patch index $k$:
$
a_{ijk}^n = \exp(s_{ijk}^n)/{\sum_{m=0}^L \exp(s_{ijm}^n)}.
$
A sentence-specific attended vision embedding is computed as the weighted sum $v_{ij}^n = \sum_{k=0}^L a_{ijk}^n v_{ik}$, which is then ℓ2-normalized as $\bar{v}_{ij}^n = v_{ij}^n / \|v_{ij}^n\|_2$. The global similarity logit between image $I_i$ and sentence $S_j^n$ is given by
$
l_{ij}^n = <{\bar{v}_{ij}^{n}}, \bar{t}_j^n> \cdot \exp(\tau).
$
The corresponding patch-level similarity map is $M_{ij}^n = [\,s_{ij1}^n, \dots, s_{ijL}^n\,]$.

\subsection{Multi-positive contrastive learning}
\label{sec:loss_func}
Although CARZero \cite{carzero}
also uses prompt templates
for training, it suffers from instability due to randomly selecting one sentence for each image at every training step. 
To utilize all \(N\) finding-sentences matched to each image at every step, we adopt 
multi-positive NCE (MP-NCE) loss \cite{uniclip} which
treats positive pairs independently in order to amplify the loss contributions from each positive pair. 
A visualization of our contrastive loss is shown in Figure \ref{fig:method} (c).
Let \(N_T=\sum_{i=1}^{B} N_i\) be the total number of finding-sentences in a mini-batch.
For the \(i\)-th image, the number of positive and negative finding-sentences are \(N_i\) and \(N_T - N_i\), respectively.
The MP-NCE loss can be computed as follows:
\begin{equation}
\mathcal{L}_I = -\frac{1}{N_T} \sum_{i=1}^{B} \sum_{n=1}^{N_i} \log \frac{\exp(l_{ii}^n)}
{\exp(l_{ii}^n) + \sum\limits_{j \neq i}^{B} \sum\limits_{m=1}^{N_j} \exp(l_{ij}^m)}
\end{equation}

For each finding-sentence \(S_{i}^n\), there is one positive image \(I_i\) and \(B - 1\) negative images. The corresponding InfoNCE loss \cite{infonce} is computed as follows:  
\begin{equation}
\mathcal{L}_T = -\frac{1}{N_T} \sum_{i=1}^{B} \sum_{n=1}^{N_i} \log \frac{\exp(l_{ii}^n)}
{\exp(l_{ii}^n) + \sum\limits_{j \neq i}^{B} \exp(l_{j,i}^n)}
\end{equation}
The final objective function is the sum of \(\mathcal{L}_I\) and \(\mathcal{L}_T\):
$
\mathcal{L} = \mathcal{L}_I + \mathcal{L}_T.
$

\subsection{Zero-shot inference}
The similarity logit between an image $I_i$ and a sentence $S_j^n$, denoted by $l_{ij}^n$, is converted into a similarity probability $\hat{l}_{ij}^n = \sigma(l_{ij}^n)$ via a sigmoid function, and used for zero-shot classification.

For grounding and segmentation, we reshape the patch-level similarity map $M_{ij}^n = [s_{ij1}^n, \dots, s_{ijL}^n]$ into a $\sqrt{L} \times \sqrt{L}$ square map, and resize it to the original image resolution via bilinear interpolation.
To account for preprocessing such as padding and resizing, this interpolation is applied accordingly.
A final element-wise sigmoid activation is applied to obtain the pixel-level similarity map $\hat{M}_{ij}^n = \sigma(\text{bilinear}(M_{ij}^n))$, which we refer to as the \textit{VL similarity map}
and use
for zero-shot grounding and segmentation.
The VL similarity map is derived from the cosine similarity between vision patches and text embeddings,
and since we do not modify the embedding space beyond applying 
ℓ2-normalization
and adjusting the temperature,
it can be directly interpreted as the similarity between each image pixel and the text.
The zero-shot inference process is illustrated in Figure \ref{fig:method} (d).

\section{Experiments}
\label{sec:experiments}

\subsection{Training dataset} \label{sec:training-data}

\paragraph{MIMIC-CXR \cite{mimic-cxr}}  
We train our model using the MIMIC-CXR dataset for VL alignment. MIMIC-CXR comprises 377,110 CXR images from 227,835 radiographic studies involving 65,379 patients. Each study includes a radiology report and one or more CXR images in either frontal or lateral views. Images are sourced from MIMIC-CXR-JPG \cite{mimic-cxr-jpg}, and only the `findings' and `impression' sections of reports are extracted using the official codebase\footnote{\url{https://github.com/MIT-LCP/mimic-cxr}}. All view positions are considered, and the official dataset split is followed. 
As described in Section~\ref{sec:finding_extraction}, finding-sentence extraction is applied, with each study containing an average of 6.45 such sentences. Studies without extracted finding-sentences are discarded, resulting in 352,875 training images and 2,852 for validation.

\subsection{Test datasets}\label{subsec:eval_dataset}

\textbf{Open-I (OI)} \cite{openi}
contains 3,851 radiology reports and 7,470 CXR images with multi-label annotations for 18 diseases.
\textbf{PadChest (PC)} \cite{padchest}
comprises 160,868 CXR images from 67,000 patients, with 192 labels showing a long-tailed distribution.
Following \cite{carzero}, we use 39,053 samples annotated by board-certified radiologists.
Additionally, \textbf{PadChest20 (PC20)}, introduced in \cite{carzero}, serves as a test set for rare disease evaluation, consisting of 20 classes with fewer than 10 samples each. 
\textbf{ChestXray14 (CXR14)} \cite{chestxray14} provides official test set with 22,433 images labeled for 14 diseases.
\textbf{CheXpert (CXP)} \cite{chexpert} includes a test set of 500 patients' images annotated by five board-certified radiologists.
Following \cite{carzero}, we evaluate classification on five observations: atelectasis, cardiomegaly, consolidation, edema, and pleural effusion. 
\textbf{ChestXDet10 (CXD10)} \cite{chestXdet10}, a subset of CXR14, contains 542 images with bounding box annotations for 10 diseases in the official test set. 
\textbf{SIIM \cite{siim}}  
pneumothorax dataset 
provides segmentation masks for 11,582 CXRs; we adopt the test split from \cite{mgca}, which includes 1,704 images with 458 positives.
\textbf{RSNA \cite{rsna}}
pneumonia dataset consists of 29,700 frontal CXRs with bounding box annotations;
we use the
test set from \cite{medklip}
containing 5,337 images, including 1,218 positives.
\textbf{MS-CXR \cite{mscxr}}
consists of 1,153 
image-phrase-bounding box
triplets, with images sourced from MIMIC-CXR.
The bounding boxes 
annotated to specific phrases in the report 
enable more detailed grounding, 
referred to as \textit{phrase grounding}.
For fair evaluation on the 
test set of 167 images
released by \cite{medrpg},
where each phrase maps to a single bounding box,
we exclude these images from the training set described in Sec. \ref{sec:training-data}.

\subsection{Evaluation metrics} \label{subsec:eval_metrics}

\textbf{AUC},
or area under the ROC curve, is adopted 
to evaluate
zero-shot classification 
on multi-label test datasets.
\textbf{Pointing game \cite{pointing}},
which determines
whether the 
coordinates of the maximum value
falls within the corresponding bounding box,
is employed as the grounding metric. 
\textbf{Dice} score serves as a standard evaluation metric for segmentation. 
Following \cite{medklip}, 
we compute the Dice score
using only positive samples and optimize the segmentation threshold
on the test set
to maximize the score. 
Threshold search intervals are 0.01 for sigmoid and 0.001 for softmax,
depending on the feature map's activation function.
\textbf{Pixel-wise AUC (Pix-AUC)}
computes AUC at pixel-level 
to evaluate the quality of the
segmentation probability map. 
To account for both sensitivity and specificity in mask prediction, we incorporate both positive and negative samples.  
For fine-grained tasks such as grounding and segmentation, predictions are interpolated back to the original image size before evaluation.

\subsection{Implementation details}
\label{sec:impl_detail}

We adopt XrayDINOv2 \cite{chexpertplus} as the pre-trained vision encoder, which was trained in a unimodal setting using CXR images based on DINOv2 \cite{dinov2}. 
While the vision encoder was trained with an image resolution of 224, we increase it to 518 for 
our experiments.
The patch size of \(14 \times 14\) leads to \(37 \times 37\) patches, yielding a vision patch length \(L\) of 1369. 
The text encoder is MPNet ("all-mpnet-base-v2") \cite{mpnet}, initialized with pre-trained parameters and further fine-tuned during training.
The trainable Transformer layers consist of two randomly initialized layers ($K=2$), with a hidden dimension of 768, matching the embedding size of both the vision and text encoders.
While the vision encoder remains frozen, all other parameters are trainable. 
Following \cite{clip}, the learnable temperature parameter $\tau$ is initialized to \(\log(1/0.07)\). 
The details of model training can be found in Appendix \ref{sec:training}.
The LLM used for extracting finding-sentences is "Llama-3.3-70B-Instruct" \cite{llama3}, deployed in a private computing environment.

\section{Results}
\label{sec:results}

\subsection{Zero-shot evaluation}\label{sec:zero-shot evaluation}

\begin{table*}[b]
\centering
\small                
\resizebox{0.9\textwidth}{!}{%
\begin{tabular}{c|c|c|c|c|c|c|c|c}
\Xhline{1pt}
\multirow{2}{*}{Method} & 
\multirow{2}{*}{\makecell{Open-I \\ (OI)}} & 
\multirow{2}{*}{\makecell{PadChest \\ (PC)}} & 
\multirow{2}{*}{\makecell{PadChest20 \\ (PC20)}} & 
\multirow{2}{*}{\makecell{ChestXray14 \\ (CXR14)}} & 
\multirow{2}{*}{\makecell{CheXpert \\ (CXP)}} & 
\multirow{2}{*}{\makecell{ChestXDet10 \\ (CXD10)}} & 
\multirow{2}{*}{SIIM} & 
\multirow{2}{*}{RSNA} \\
& & & & & & & & \\
\hline\hline
GLoRIA \cite{gloria} & 0.589  & 0.565    & 0.558      & 0.610       & 0.750    & 0.645       & - & - \\
BioViL-T \cite{biovil-t} & 0.702  & 0.655    & 0.608      & 0.729       & 0.789    & 0.708       & - & - \\
MedKLIP \cite{medklip} & 0.759  & 0.629    & 0.688      & 0.726       & 0.879    & 0.713       & 0.897 & \textbf{0.869} \\
KAD     \cite{kad}     & 0.807  & 0.750    & 0.735      & 0.789       & 0.905    & 0.735       & -     & -     \\
CARZero \cite{carzero} & 0.838  & 0.810    & 0.837      & \textbf{0.811}       & \textbf{0.923}    & \textbf{0.796}       & \textbf{0.924} & 0.747 \\
RadZero~(224px)
& \textbf{0.851}  & \textbf{0.841}    & \textbf{0.879}      & 0.807       & 0.903    & 0.785       & 0.914 & 0.839 \\
RadZero   & 0.847  & \textbf{0.841}    & 0.871      & 0.804       & 0.900    & 0.787       & \textbf{0.924} & 0.834 \\
                               
\Xhline{1pt}
\end{tabular}}
\caption{
Zero-shot classification AUROC scores on public CXR datasets.
For fair comparison, we also report the results of low-resolution (224×224) version of RadZero.
}
\label{tab:zero-shot-cls}
\end{table*}

\paragraph{Classification.}
Table \ref{tab:zero-shot-cls} compares
RadZero with SOTA models on public test datasets.
For the five datasets evaluated in CARZero \cite{carzero}, 
we report their published results.
For SIIM and RSNA, we independently evaluated two open-source models.
RadZero achieved new SOTA performance on 
OI and PC,
irrespective of image resolution.
In the long-tailed PC dataset
with 192 classes, RadZero outperformed 
CARZero
by 3.1 
percentage
points, 
demonstrating strong generalization in zero-shot classification.
Notable gains are also observed in PC20,
which focuses on rare diseases, 
suggesting that 
VL-CABS
is particularly effective for infrequent conditions.
On datasets where RadZero failed to rank first,
MedKLIP performed best on RSNA, while CARZero led on 
CXR14, CXP, and CXD10.
However, MedKLIP underperformed on the latter datasets, and CARZero underperformed on RSNA.
In contrast, RadZero showed results comparable to the top-performing models across all datasets.
Interestingly, the lower-resolution RadZero (224px) even outperformed RadZero:
potentially due to the pre-trained vision encoder,
as discussed in Sec. \ref{sec:ablation}.

The representative classification metric shown in Figure \ref{fig:introduction-performance} is the average AUC across all datasets.
RadZero established a new SOTA,
outperforming CARZero by 1.5 percentage points, 
a gain attributable to our training strategy
that
incorporates multi-positive 
contrastive learning
to enhance the diversity of both positive and negative samples per image.

\paragraph{Grounding.}

\begin{table*}[t]
\centering
\small                
\resizebox{0.9\textwidth}{!}{
\begin{tabular}{c|c|c|c|c|c|c|c|c|c|c|c}
\Xhline{1pt}
Method  & Mean & ATE & CALC & CONS & EFF & EMPH & FIB & FX & MASS & NOD & PTX \\
\hline\hline
GLoRIA   \cite{gloria}   & 0.367 &	0.479 & 0.053 &	0.737 &	0.528 &	0.667 &	0.366 &	0.013 &	0.533 &	0.156 &	0.143 \\
KAD      \cite{kad}      & 0.391 &	\textbf{0.646} & 0.132 &	0.699 &	0.618 &	0.644 &	0.244 &	0.199 &	0.267 &	0.316 &	0.143 \\
BioViL-T \cite{biovil-t} & 0.351 &  0.438 & 0.000 & 0.630 & 0.504 & 0.846 & 0.390 & 0.026 & 0.500 & 0.000 & 0.171\\
MedKLIP  \cite{medklip}  & 0.481 &	0.625 & 0.132 &	\textbf{0.837} &	0.675 &	0.734 &	0.305 &	0.224 &	0.733 &	0.312 &	0.229 \\
CARZero  \cite{carzero}  & 0.543 &	0.604 & 0.184 &	0.824 &	0.782 &	0.846 &	0.561 &	0.184 &	0.700 &	0.286 &	0.457 \\
RadZero~(224px)
& 0.537 &  0.604 & 0.211 & 0.806 & 0.813 & 0.795 & 0.451 & 0.197 & \textbf{0.767} & 0.325 & 0.400       \\                     
RadZero                  & \textbf{0.622} &  \textbf{0.646} & \textbf{0.368} & 0.824 & \textbf{0.857} & \textbf{0.872} & \textbf{0.585} & \textbf{0.250} & \textbf{0.767} & \textbf{0.506} & \textbf{0.543}       \\                     
\Xhline{1pt}
\end{tabular}}
\caption{
Zero-shot grounding results (pointing game accuracy) on CXD10.
Lesion abbreviations can be found in Appendix \ref{appendix:abbreviation}.
}
\label{tab:ground_findings}
\end{table*}

\begin{wraptable}{r}{0.3\textwidth}  
\centering
\small           
\resizebox{0.3\textwidth}{!}{%
\begin{tabular}{c|c}
\Xhline{1pt}
Method  & MS-CXR \\
\hline\hline
BioViL-T \cite{biovil-t} & 0.719 \\
MedKLIP  \cite{medklip}  & 0.407 \\
CARZero  \cite{carzero}  & 0.749 \\
RadZero~(224px) & 0.832 \\   
RadZero  & \textbf{0.844} \\
\Xhline{1pt}
\end{tabular}
}
\caption{
Zero-shot phrase grounding results (pointing game accuracy) on MS-CXR.
}
\label{tab:zero-shot-grnd}
\end{wraptable}

Table \ref{tab:ground_findings} presents zero-shot grounding results on 
CXD10.
We adopted the pointing game scores 
reported by CARZero
for all models except BioViL-T,
which we evaluated using its released weights.
RadZero achieved the highest average score across all diseases, outperforming CARZero by 0.079.
Per-lesion analysis showed that 
RadZero achieved the best performance in all classes except consolidation, 
indicating that the proposed 
VL-CABS
effectively captures local alignment between text and image patches regardless of disease type.
Furthermore, RadZero efficiently supports higher input resolutions, enabling more precise localization.

Table \ref{tab:zero-shot-grnd} reports zero-shot phrase grounding results, evaluating alignment at the phrase level in contrast to disease-level grounding in Table \ref{tab:ground_findings}.
Pointing game accuracy is used as the evaluation metric.
We evaluated all baselines using publicly available models. 
RadZero achieved the highest score of 0.844, 
demonstrating accurate interpretation of text phrases.
The strong performance of RadZero (224px) suggests that the gains are largely attributable to the effectiveness of VL-CABS rather than the higher input resolution.

\paragraph{Segmentation.}

\begin{wraptable}{r}{0.4\textwidth}  
  \centering
  \small              
  \resizebox{0.4\textwidth}{!}{%
\begin{tabular}{c|c|c|c}
\Xhline{1pt}
\multirow{2}{*}{Method}  & RSNA & \multicolumn{2}{c}{SIIM} \\
\cline{2-4}
                         & Dice & Dice & Pix-AUC \\
\hline\hline
GLoRIA  \cite{gloria}  & 0.347\textsuperscript{*} & -  & -    \\
BioViL  \cite{biovil}  & 0.439\textsuperscript{*} & -  & -    \\
MedKLIP \cite{medklip} & 0.465\textsuperscript{*} & 0.044 & 0.648\\
G2D     \cite{g2d}     & 0.477$^\dagger$ & 0.051$^\dagger$ & -\\
CARZero \cite{carzero} & 0.540 & 0.100 & 0.856 \\   
CARZero (logits)       & 0.529 & 0.081 & 0.928 \\ 
RadZero~(224px) & \textbf{0.562} & 0.121 & 0.943 \\
RadZero  & 0.546 & \textbf{0.171} & \textbf{0.947} \\                             
\Xhline{1pt}
MGCA \cite{mgca} (1\%)   & 0.513 & 0.144 & 0.752 \\
MGCA (10\%)              & 0.571 & 0.238 & 0.856 \\
MGCA (100\%)             & 0.578 & 0.305 & 0.976 \\
\Xhline{1pt}
\end{tabular}
}
\caption{
\footnotesize
Zero-shot segmentation results.
Values with $^*$ are from \cite{medklip} and $^\dagger$ from \cite{g2d}
}
\label{tab:zero-shot-seg}
\end{wraptable}

Table \ref{tab:zero-shot-seg} summarizes zero-shot segmentation results on SIIM and RSNA. 
To benchmark against supervised models, we fine-tuned MGCA with varying proportions of training data; percentages in parentheses indicate the amount used.

Among zero-shot models, RadZero achieved the highest Dice scores on both datasets. 
On SIIM, it outperformed CARZero by 71\%,
demonstrating superior segmentation capability. 
The smaller margin on RSNA is likely due to its coarser annotations—bounding boxes rather than pixel-level masks—which limit the advantages of RadZero’s fine-grained VL similarity maps.
As in phrase grounding, the 
results
of RadZero (224px) suggest that the superior performance is not merely driven by higher resolution.

RadZero also remained competitive against 
fine-tuned models.
It outperformed MGCA (1\%) on both datasets, demonstrating the 
effectiveness 
of zero-shot segmentation.
Although it did not surpass
MGCA (10\%) or (100\%), 
RadZero requires no mask labels,
enabling broader generalization beyond fixed vocabularies,
as further discussed in Sec. \ref{sec:ov-seg}.

SIIM’s detailed annotations also allow for Pix-AUC evaluation.
RadZero achieved the highest 
score, exceeding 
even MGCA (10\%), 
indicating
well-calibrated VL similarity maps that 
distinguish positive from negative regions.
In contrast, MedKLIP and CARZero, 
both relying 
on attention maps, 
performed worse with 
scores
of 0.648 and 0.856, respectively.
For a fair comparison, 
we also evaluated CARZero's pre-softmax logits
(CARZero (logits)), which improved performance but still fell short of RadZero.
Notably, 
CARZero (logits) 
underperformed its own
Dice score (0.081 vs. 0.100) 
despite threshold tuning,
as expected from the inconsistent scaling of dot-product similarity.
In contrast, 
VL-CABS
directly encode pixel-level 
text–image 
similarity, allowing low
values 
for negative samples.
This contributed to its superior Pix-AUC,
as further supported 
in Sec. \ref{sec:sim-map}.

\begin{figure}[t]
  \centering
  \begin{minipage}[t]{0.48\textwidth}
    \centering
    \includegraphics[width=\linewidth]{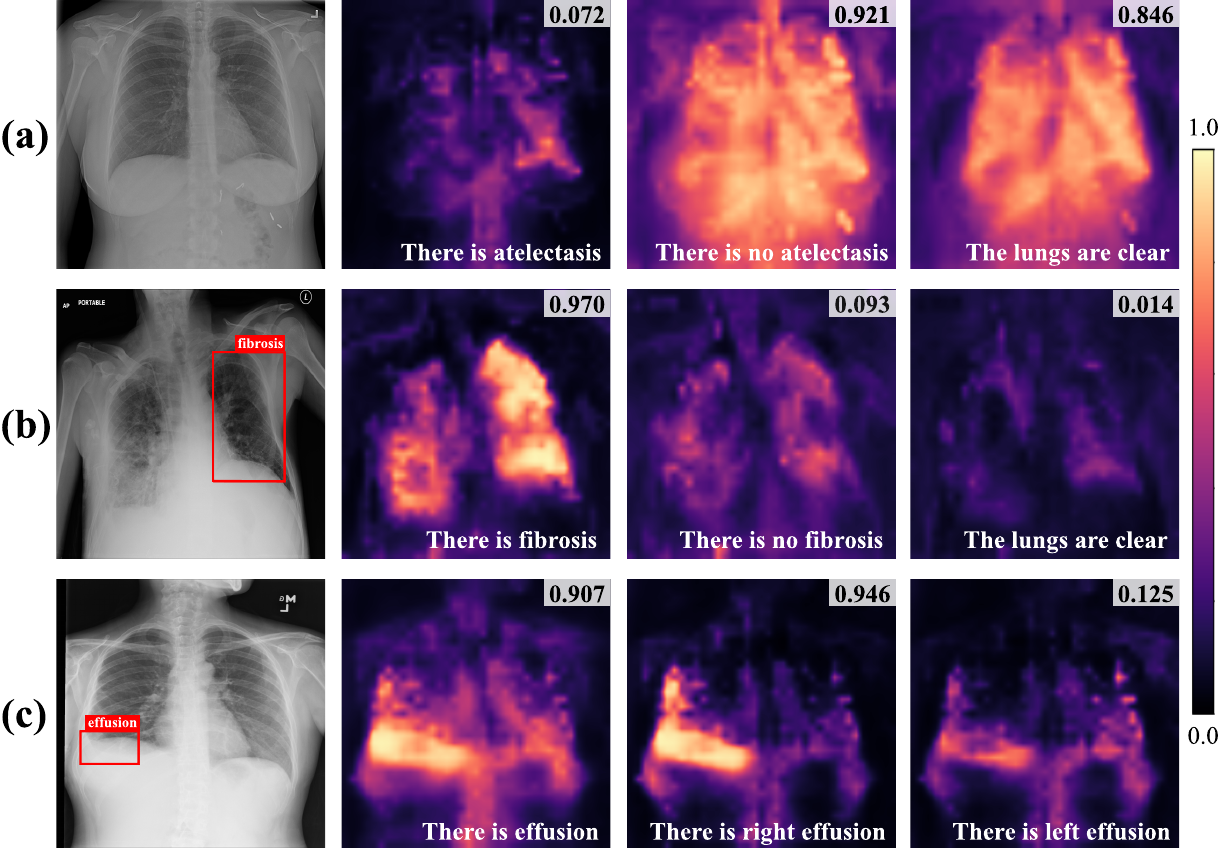}
    \caption{
    VL similarity maps of CXR images from CXD10,
    representing
    (a) normal, (b) fibrosis, and (c) effusion in the right lung.
    The value at the top-right corner represent
    the similarity probability \(\hat{l}\)
    between each 
    CXR image and the text prompt (bottom-right corner). 
    }
    \label{fig:segmentation probability}
    \end{minipage}%
  \hfill
  \begin{minipage}[t]{0.48\textwidth}
    \centering
    \includegraphics[width=\linewidth]{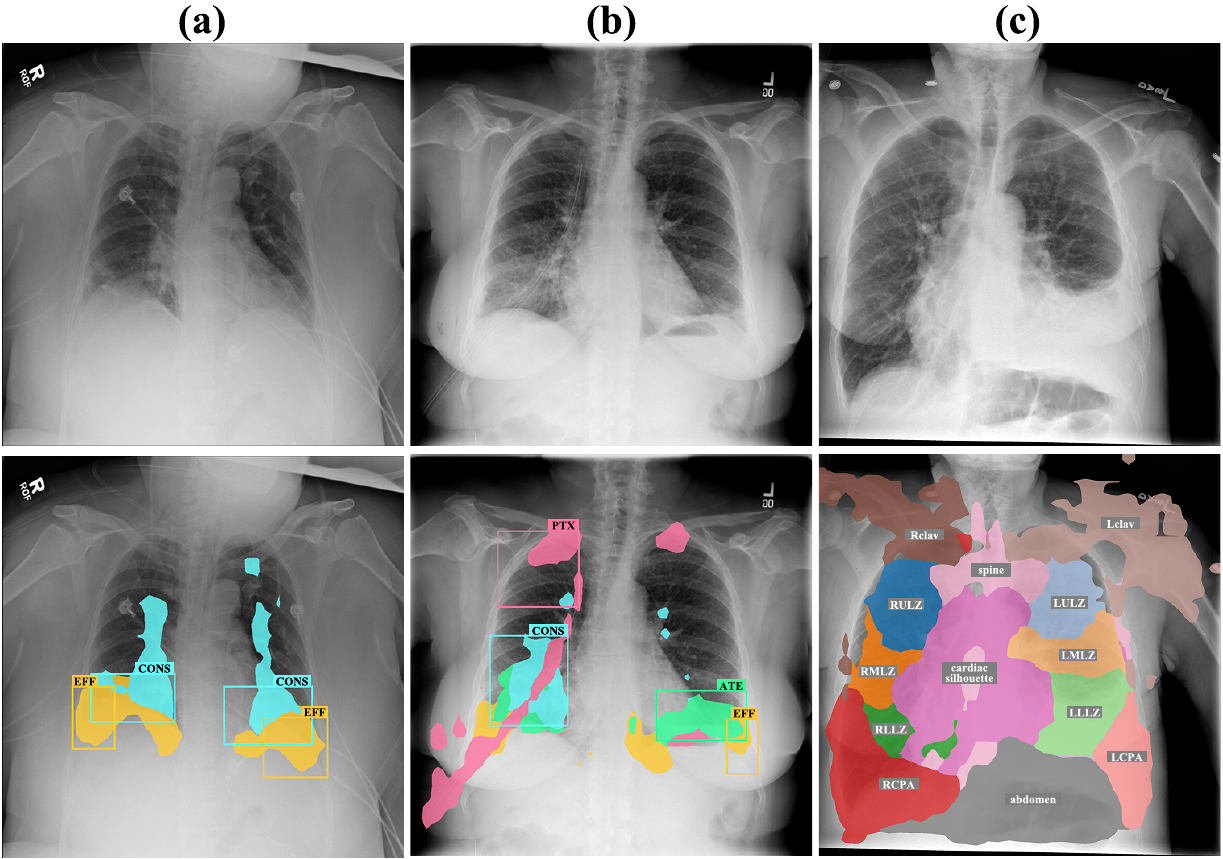}
    \caption{
    Open-vocabulary semantic segmentation results: (a), (b) for findings and (c) for anatomical regions. 
    The CXR images and bounding box labels are from CXD10. 
    The segmentation thresholds were set to 0.7 for (a) and (b), and 0.4 for (c).
    }
    \label{fig:findings segmentation}  
  \end{minipage}
\end{figure}

\subsection{VL similarity map analysis}
\label{sec:sim-map}

Figure \ref{fig:segmentation probability} illustrates that
RadZero effectively aligned
visual and textual representations 
with VL-CABS.
The outputs, 
VL similarity map \(\hat{M}\)
and probability \(\hat{l}\),
offer both interpretable visualizations and quantitative metrics.
For the normal image Figure \ref{fig:segmentation probability} (a), 
the model assigned low similarity (0.072) to the prompt ``There is atelectasis''
with a dark VL similarity map, indicating 
weak alignment between vision tokens and the text embedding, 
In contrast, ``There is no atelectasis'' (0.921) and ``The lungs are clear'' (0.846) yielded 
bright activations across lung fields, reflecting strong alignment.

Figure \ref{fig:segmentation probability} (b) shows fibrosis, and 
``There is fibrosis'' 
resulted in high similarity (0.970)
with strong activations in the affected lung.
Prompts indicating normality received 
much lower scores 
(0.093 and 0.014)
and darker VL similarity maps,
clearly distinguishing abnormal from normal descriptions.

Figure \ref{fig:segmentation probability} (c) highlights RadZero’s ability to 
distinguish anatomical descriptions.
For right-sided pleural effusion, 
the model assigned high similarity (0.907) to ``There is effusion,'' with bright activations in the correct region.
Notably, ``There is right effusion'' (0.946) scored even higher, indicating accurate localization, while ``There is left effusion'' scored much lower (0.125) and a dark VL similarity map, 
showing that the model correctly distinguishes between left and right lung regions.

Overall, these results underscore 
the explainability of VL-CABS. 
The similarity probability is verifiable at the pixel level, 
enabling spatially grounded explanations. 
By explicitly revealing how conclusions are derived, 
RadZero offers enhanced interpretability in the context of disease diagnosis.

\subsection{Open-vocabulary semantic segmentation} 
\label{sec:ov-seg}

Figure \ref{fig:findings segmentation} presents open-vocabulary semantic segmentation results for both
findings and anatomical regions.
Segmentation masks were generated by thresholding the VL similarity map \(\hat{M}\)
for each text prompt;
in cases of overlapping predictions, 
the prompt with the highest similarity was assigned.
In Figure \ref{fig:findings segmentation} (a), RadZero
successfully localized lesions based on text queries,
though some segmentation masks extended beyond ground truth boxes, 
indicating room for improvement.
Notably, certain 
incorrect predictions
captured clinically relevant features that were 
not explicitly annotated:
in Figure \ref{fig:findings segmentation} (b),
a chest tube was reasonably associated with ``pneumothorax.''
Figure \ref{fig:findings segmentation} (c) further demonstrates RadZero's ability to
segment anatomical structures without supervision, 
inferring approximate spatial regions from text despite imprecise boundaries. 
These results highlight 
the potential of VL-CABS
for zero-shot open-vocabulary semantic segmentation 
and RadZero's capacity to align textual descriptions with medical imagery.
Additional qualitative examples are provided in Appendix~\ref{sec:additional_visualization}.

\subsection{Ablation Studies}
\label{sec:ablation}
\begin{table*}[!t]
  \centering
  \small                
  \resizebox{\textwidth}{!}{
    \begin{tabular}{l cc c c c c c c c c c c c}
      \toprule
      \multirow{2}{*}{Method} & \multirow{2}{*}{Similarity} & \multirow{2}{*}
      {\makecell{Trainable \\ layers}} & 
      \multirow{2}{*}{MP} & \multirow{2}{*}{Res.} 
            & \multicolumn{6}{c}{Classification} 
            & \multicolumn{2}{c}{Grounding} 
            & Segmentation \\
      \cmidrule(lr){6-11} \cmidrule(lr){12-13} \cmidrule(lr){14-14}
       &       &   &    &    
            & OI & PC & PC20 & CXR14 & CXP & CXD10 
            & CXD10 & MS-CXR & SIIM \\
      \midrule

      (a)           & dot-product & Linear    & 
      \ding{55}  & 224 & 0.839 & 0.824 & 0.853 & 0.805 & 0.896 & 0.792 & 0.472 & 0.784 & 0.078 \\
      (b)          & $\cos$ & Linear    & 
      \ding{55}  & 224 & 0.843 & 0.830 & 0.863 & 0.805 & 0.902 & 0.786 & 0.483 & 0.790 & 0.078 \\
      (c)           & $\cos$ & 2 Transformer  & 
      \ding{55}  & 224 & 0.845 & 0.832 & 0.860 & \textbf{0.808} & 0.895 & \textbf{0.793} & 0.539 & 0.838 & 0.099 \\
      (d) RadZero~(224px) & $\cos$ & 
      2 Transformer  & \ding{51}  & 224 & 
      \textbf{0.851} & \textbf{0.841} & \textbf{0.879} & 0.807 & \textbf{0.903} & 0.785 & 0.537 & 0.832 & 0.121 \\
      \textbf{RadZero}       & $\cos$ & 
      2 Transformer  & \ding{51} & 518 & 0.847
      & \textbf{0.841} & 0.871 & 0.804 & 0.900 & 0.787 & \textbf{0.622} & \textbf{0.844} & \textbf{0.171} \\
   
      \midrule

      LiT \cite{lit}           & - & Linear    & 
      \ding{55}  & 224 & 0.768 & 0.769 & 0.775 & 0.764 & 0.854 & 0.735 & -     & -    & - \\
      \texttt{dino.txt} \cite{dinotxt}     & - & 
      2 Transformer  & \ding{55} & 224 & 0.834
      & 0.816 & 0.837 & 0.797 & 0.901 & 0.770 & 0.121 & 0.174 & 0.021 \\
      CARZero \cite{carzero}       & - & 
      Transformer Dec.   & \ding{55}  & 224 & 
      0.827 & 0.815 & 0.877 & 0.795 & 0.889 & 0.770 & 0.437 & 0.743 & 0.072 \\
      
      \bottomrule
    \end{tabular}
  }
  \caption{Ablation study of model architecture components. 
  ``MP'' denotes 
  multi-positive.
  }
  \label{tab:ablation}
\end{table*}

\paragraph{Ablation study on RadZero components.}
\textbf{1) Similarity function:}
Comparing (a) and (b) in Table~\ref{tab:ablation} shows the effect of using cosine similarity instead of 
scaled dot-product \cite{transformer} for
VL alignment.
Cosine similarity, which better aligns with inference-time VL similarity maps, improved classification and slightly enhances grounding.
\textbf{2) Trainable parameters:}
(b) and (c) compare a linear layer and a two-layer Transformer, with the image encoder frozen.
Transformer layers
yielded consistent gains across all tasks, particularly in grounding (0.483 → 0.539) and segmentation (0.078 → 0.099).
\textbf{3) Multi-positive pairs:}
The difference between (c) and (d) lay in the use of multi-positive contrastive pairs.
(d) improved classification 
(e.g., PC20: 0.860 → 0.879) 
and segmentation (0.099 → 0.121), highlighting the advantage of richer supervision.
\textbf{4) Image resolution:}
RadZero and (d) shared the same architecture, except that RadZero used higher resolution inputs (518 vs. 224).
This change substantially improved 
grounding (0.537 → 0.622) and segmentation (0.121 → 0.171), 
showing the importance of high-resolution features for spatially localized tasks.
In classification, (d) outperformed RadZero: likely due to the vision encoder (XrayDINOv2\cite{chexpertplus}) being pre-trained at 224 pixels,
indicating that using an encoder pre-trained at higher resolutions may further enhance the performance.

\paragraph{Comparison among different VL alignment approaches.}
Table~\ref{tab:ablation} compares RadZero with alternative VL alignment methods,
keeping all settings identical except for VL feature fusion and loss computation.
LiT, which uses a \texttt{[CLS]} embedding for alignment, 
showed limited classification performance
and was incapable of 
grounding or segmentation. 
\texttt{dino.txt} improved classification through additional Transformer layers, 
but its mean pooling constrained grounding and segmentation performance. 
CARZero introduced a cross-attention decoder, enhancing performance on those tasks. 
However, when compared to the RadZero ablations, 
(b) outperformed CARZero across most metrics,
showing that VL-CABS alone is sufficiently effective.

\section{Conclusion}
\label{sec:conclusion}

In this work, we introduced RadZero, a novel 
VL alignment model
for chest X-ray 
that achieved strong zero-shot performance in classification, grounding, and segmentation. 
Central to RadZero is VL-CABS, 
which computes 
image-text
similarity at the patch-level to improve interpretability. 
Combined with multi-positive contrastive training, 
VL-CABS
enabled effective representation learning without 
pixel-level annotations,
and the support for high-resolution inputs 
further boosted performance. 
Extensive evaluations on public chest radiograph benchmarks 
showed that RadZero outperformed
SOTA methods.
VL similarity map analysis highlighted the
enhanced explainability of VL-CABS
by providing transparent rationales for 
how conclusions are derived.
Qualitative assessments further demonstrated
RadZero’s potential for open-vocabulary semantic segmentation.

Despite its impressive results, RadZero
has limitations that 
indicate areas for future research.
The observed performance degradation on specific datasets emphasizes the necessity of enhancing generalization capability.
Its reliance on the pre-trained vision encoder may also restrict domain adaptability.
In addition, the current study validates the proposed RadZero training framework only on chest X-ray datasets, which limits the scope of its generalization.
Future work could explore extending RadZero to other medical imaging modalities such as CT and MRI, demonstrating its potential as a universal vision-language learning framework adaptable to diverse anatomical and visual characteristics.
Moreover, applying VL-CABS to general imaging domains, for instance in open-vocabulary semantic segmentation, could be a meaningful direction toward building more interpretable VLMs.

\begin{ack}
This work was supported by the Technology Innovation Program (RS-2025-02221011, Development of Medical-Specialized Multimodal Hyperscale Generative AI Technology for Global Integration) funded by the Ministry of Trade Industry \& Energy (MOTIE, South Korea).
\end{ack}

{
\small
\bibliographystyle{plainnat}
\bibliography{main}
}

\clearpage

\appendix

\section{Are Attention Maps of Medical VLMs Fully Explainable?}
\label{appendix:attenttion_map}

Explainability is critical for clinical deployment of deep learning models, and medical VLMs have adopted attention maps as their de facto interpretable features \cite{medklip, carzero, kad}. 
However, attention maps alone reveal only \textit{where} the model 
is focusing, not \textit{why} it makes a particular prediction.
Without the underlying image–text similarity scores, such heatmaps lack interpretability. 
Interpreting attention maps for complex text queries can be unintuitive and may depend on access to ground-truth labels.

\begin{wrapfigure}{r}{0.45\textwidth}
    \centering
    \includegraphics[width=\linewidth]{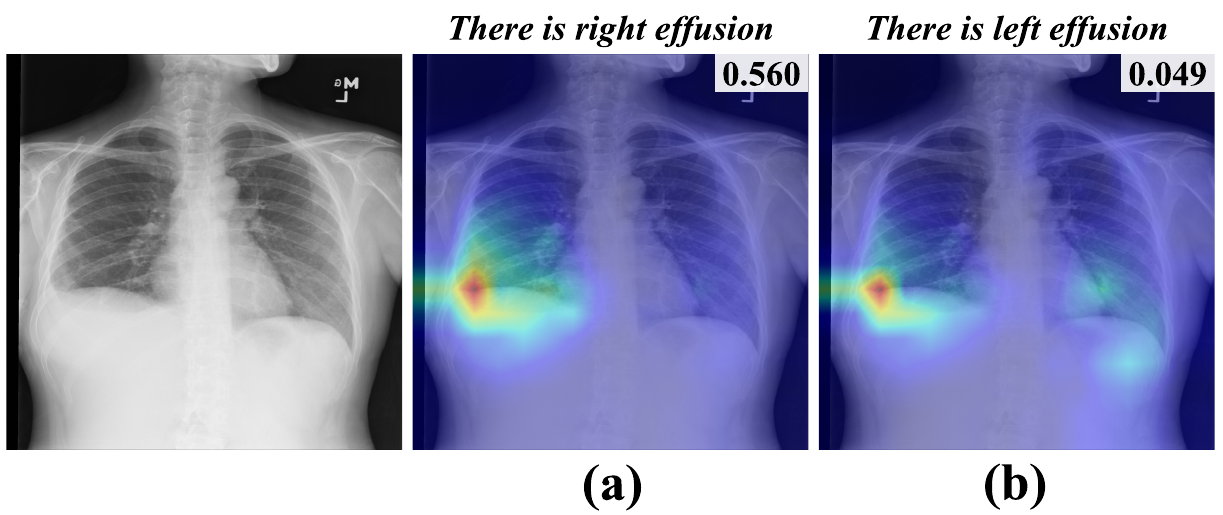}
    \caption{
        Attention maps from CARZero \cite{carzero}. Image–text similarity probabilities are obtained by applying a sigmoid function to the classification logits.
    }
    \label{fig:carzero_attention_map}
\end{wrapfigure}

To illustrate the challenges of interpreting attention maps, Figure \ref{fig:carzero_attention_map} visualizes the attention outputs of CARZero \cite{carzero} for different image–text pairs, with the overlaid values indicating the corresponding similarity probabilities.
In (a), the prompt ``There is right effusion'' produces a heatmap focused on the right lower lung and a high similarity probability (0.560), intuitively linking the attended region to the predicted finding. 
In (b), when queried with “There is left effusion” on the same image, the attention map still highlights the right lower lung. 
While we can observe \textit{where} the model attends, the attention map alone offers no clear explanation for \textit{why} that region is relevant. 
The image–text similarity probability for the prompt “There is left effusion” is low (0.048), which—when considered alongside the attention map—can be interpreted as indicating that the attended right lung region does not support the presence of left-sided effusion.
However, how should we interpret the fact that the model attends to the right lung when queried about \textit{left} effusion? Making sense of this behavior requires access to the ground truth: effusion is present in the right lower lung but absent on the left. 
With this context, one might infer that the model correctly identifies effusion on the right, recognizes that it does not match the left-sided query, and therefore assigns a low similarity score.

This example underscores the need to interpret attention maps in conjunction with image–text similarity scores. 
For complex text queries, meaningful interpretation often depends on knowing the ground truth—a significant limitation in medical VLMs where such labels are frequently unavailable. 
This reliance undermines the standalone explainability of attention visualizations. 
To address these limitations, we propose vision–language cross-attention based on similarity (VL-CABS), a framework that enables transparent inspection of the vision–language decision process.
Detailed examples and analysis of its application in RadZero are provided in Sec. \ref{sec:sim-map}.

\section{Ablation Studies} \label{appendix:ablation_study}

We conducted ablation studies to assess the impact of key design choices by selectively modifying parts of our approach.
The zero-shot tasks for evaluation included 
classification (\textit{class.}), 
grounding (\textit{ground.}), 
phrase grounding (\textit{phrase.}), 
and segmentation (\textit{seg.}).
Classification was tested on the PadChest dataset, known for its highly imbalanced (long-tailed) label distribution, with AUC as the evaluation metric.  
Grounding and phrase grounding were evaluated using the pointing game on the ChestXDet10 and MS-CXR test sets, respectively.  
Segmentation performance was measured by the Dice score on the SIIM dataset.  
For the ablation study, the default batch size and maximum number of epochs were set to 128 and 10, respectively.

\paragraph{View position.}

\begin{wraptable}{r}{0.48\textwidth}  
    \centering
\scalebox{.85}{
\begin{tabular}{c|c|c|c|c}
\Xhline{1pt}
View Position  & \textit{class.} & \textit{ground.} & \textit{phrase.}  & \textit{seg.} \\
\hline\hline
Frontal            & 0.831 & 0.604 & 0.838 & 0.161     \\
\textbf{All View}  & \textbf{0.841} & \textbf{0.622} & \textbf{0.844} & \textbf{0.171}     \\
\Xhline{1pt}

\end{tabular}}

\caption{Impact of view position.} 
\label{tab:apx_view}

\end{wraptable}

Table \ref{tab:apx_view} shows the performance variation based on the view position of CXR images. We compared two models: one trained exclusively on frontal view images from MIMIC-CXR and another trained on both frontal and lateral views. The model trained on all view positions consistently outperformed the frontal-only model, suggesting that it effectively learned to interpret lateral images, enhancing overall robustness.

\paragraph{Vision encoder and resolution.}

Table \ref{tab:apx_img_encoder} presents the impact of the vision encoder and image resolution on model performance. We compared 
DINOv2 \cite{dinov2},
RadDINO \cite{raddino} and XrayDINOv2 \cite{chexpertplus} using image resolutions of 224 and 518. 
DINOv2, which was trained on natural images rather than X-rays, exhibited relatively lower performance, as expected due to the domain mismatch.
Comparing XrayDINOv2 at resolutions of 224 and 518, we observe that higher image resolution improves fine-grained tasks such as grounding and segmentation.  
RadDINO and XrayDINOv2 showed similar performance, suggesting that our approach is effectively applied to models trained with the DINOv2 strategy on chest X-ray images.

\paragraph{Trainable vision layer architecture.}

\begin{table*}[t]
\centering

\begin{minipage}[t]{0.48\textwidth}
    \centering

\centering
\scalebox{.65}{
\begin{tabular}{c|c|c|c|c|c}
\Xhline{1pt}
Vision   & Image  & \multirow{2}{*}{\textit{class.}} & \multirow{2}{*}{\textit{ground.}} & \multirow{2}{*}{\textit{phrase.}}  & \multirow{2}{*}{\textit{seg.}} \\
Encoder & Resolution & & & & \\
\hline\hline
DINOv2 \cite{dinov2} & 518   & 0.825 & 0.606 & 0.814 & 0.100 \\
RadDINO \cite{raddino} & 518   & \textbf{0.850} & 0.610 & \textbf{0.844} & 0.144 \\
XrayDINOv2 \cite{chexpertplus} & 224   & 0.841 & 0.548 & 0.832 & 0.118 \\
\textbf{XrayDINOv2} & 518   & 0.841 & \textbf{0.622} & \textbf{0.844} & \textbf{0.171} \\
\Xhline{1pt}
\end{tabular}
}

\caption{Impact of vision encoder and resolution.}
\label{tab:apx_img_encoder}
\end{minipage}
\hfill
\begin{minipage}[t]{0.48\textwidth}
    \centering
    \centering
\scalebox{.8}{
\begin{tabular}{c|c|c|c|c}
\Xhline{1pt}
Model  & \textit{class.} & \textit{ground.} & \textit{phrase.}  & \textit{seg.} \\
\hline\hline
Linear                       & 0.826 & 0.549 & 0.826 & 0.100      \\
1 Transformer layer          & 0.835 & 0.585 & 0.832 & 0.158      \\
\textbf{2 Transformer layers} & \textbf{0.841} & \textbf{0.622} & \textbf{0.844} & \textbf{0.171}      \\
\Xhline{1pt}

\end{tabular}}

\caption{Impact of trainable vision layer.}
\label{tab:apx_trainable_layer}

\end{minipage}
\begin{minipage}[t]{0.48\textwidth}
    \centering
    \centering
\scalebox{.85}{
\begin{tabular}{c|c|c|c|c}
\Xhline{1pt}
Text   & \multirow{2}{*}{\textit{class.}} & \multirow{2}{*}{\textit{ground.}} & \multirow{2}{*}{\textit{phrase.}}  & \multirow{2}{*}{\textit{seg.}} \\
Encoder & & & & \\
\hline\hline
BioBERT             & \textbf{0.842} & 0.582 & 0.832 & 0.127      \\
\textbf{MPNet} \cite{mpnet}      & 0.841 & \textbf{0.622} & \textbf{0.844} & \textbf{0.171}      \\
\Xhline{1pt}

\end{tabular}}

\caption{Impact of text encoder.}
\label{tab:apx_txt_encoder}

\end{minipage}
\hfill
\begin{minipage}{0.48\textwidth}
    \centering
    \centering
\scalebox{.85}{
\begin{tabular}{c|c|c|c|c}
\Xhline{1pt}
Batch size  & \textit{class.} & \textit{ground.} & \textit{phrase.}  & \textit{seg.} \\
\hline\hline
64              & 0.835 & 0.583 & 0.826 & 0.165      \\
128             & 0.840 & 0.594 & \textbf{0.850} & \textbf{0.177}      \\
\textbf{256}    & \textbf{0.841} & \textbf{0.622} & 0.844 & 0.171     \\
\Xhline{1pt}

\end{tabular}}

\caption{Impact of batch size.}
\label{tab:apx_batch}

\end{minipage}

\end{table*}

Table \ref{tab:apx_trainable_layer} presents the impact of different trainable layers in the image encoder. The commonly used linear layer showed relatively lower performance across tasks. In contrast, two Transformer layers achieved the best results 
across all tasks.
Based on this observation, RadZero was designed with two Transformer layers added to the vision encoder. This improvement is likely due to the Transformer's ability to attend to all patch embeddings, capturing richer semantic information.

\paragraph{Text encoder.}

Table \ref{tab:apx_txt_encoder} presents the performance of different text encoders used during training. We compared MPNet \cite{mpnet} and BioBERT \cite{biobert}, where BioBERT was fine-tuned on clinical reports by CARZero \cite{carzero}.  
While MPNet showed slightly lower performance in classification, it achieved notable improvements in phrase grounding and segmentation, demonstrating its effectiveness in tasks requiring fine-grained text-image alignment.

\paragraph{Batch size.}

Table \ref{tab:apx_batch} presents the impact of batch size on model performance during training. To ensure a fair comparison, we maintained a consistent total number of training steps by adjusting the number of epochs: 5 for a batch size of 64, 10 for 128, and 20 for 256.  
We observed that a batch size of 64 resulted in lower performance across all tasks. While the model trained with a batch size of 128 performed reasonably well, its zero-shot grounding performance was notably lower than that of the 256 batch size model. As a result, we selected 256 as the final batch size.  
This trend aligns with the well-known impact of batch size in contrastive learning, where larger batch sizes generally improve representation learning by providing more diverse negative samples, leading to better alignment and discrimination.

\section{Model Training and Computational Details}
\subsection{Training Configuration} \label{sec:training}
RadZero is trained for 20 epochs with an early stopping patience of 5 epochs, selecting the best model based on validation loss. 
We employ the AdamW optimizer with a learning rate of 0.0001, following a cosine decay scheduler, with 50 warm-up steps, a weight decay of 0.05, and gradient clipping set to 1.0. 
Training is conducted with a global batch size of 256 using distributed data parallel (DDP) on four Nvidia H100 GPUs for 13 hours. 

\subsection{Resolution Trade-off Analysis}
\label{appendix:resolution-tradeoff}
RadZero employs high-resolution images of 518 px instead of 224 px, 
which naturally increases the computational cost. 
Table \ref{tab:apx_training_cost} summarizes the trade-off between performance and resource usage across different image resolutions.
The columns with downward arrows ($\downarrow$) indicate that lower values are better.

\begin{table*}[h]
\centering
\small                
\resizebox{1.0\textwidth}{!}{
\begin{tabular}{l|c|c|c|c|c|c|c|c}
\Xhline{1pt}
\multirow{2}{*}{Method} 
& \multicolumn{2}{c|}{\textbf{GPU Memory}$\downarrow$ (GB)}
& \textbf{Training}
& \textbf{Latency}$\downarrow$
& \textbf{Throughput}
& \textbf{Cls} 
& \textbf{Grnd} 
& \textbf{Seg} \\
\cline{2-3}
& \multicolumn{1}{c|}{Training} & \multicolumn{1}{c|}{Inference} &  \textbf{GPU hour}$\downarrow$ & (ms/img) & (img/s) & \textbf{AUC} & \textbf{ACC} & \textbf{DICE} \\
\hline
RadZero~(224px) & 25.13 $\times$ 4 & 2.20 & 40 & 60.18 & 733.56 & 0.852 & 0.537 & 0.342 \\
RadZero & 73.36 $\times$ 4 & 2.38 & 52 & 70.05 & 94.93 & 0.851 & 0.622 & 0.359 \\
\Xhline{1pt}
\end{tabular}}
\caption{Performance–cost trade-off across different image resolutions.}
\label{tab:apx_training_cost}
\end{table*}
Both models were trained using four GPUs with a batch size of 64 per device (total batch size of 256). 
For inference, memory usage and latency were measured with a batch size of 1. 
Throughput (images per second) was measured using the largest power-of-two batch size that fits into GPU memory for each model: 4096 for the 224 px model and 256 for the 518 px model. 
Note that throughput (img/s) and latency (ms/img) are not exact reciprocals because throughput is measured under a large-batch setting where computation is parallelized across samples, while latency reflects the time required to process a single image without such parallelism. 
All computational cost measurements were conducted on Nvidia H100 GPUs using the ChestXray14 dataset.
The reported performance represents the average over all datasets for each task.
The results indicate that the additional computational cost is a worthwhile trade-off given the substantial improvement in fine-grained performance.

\section{Detailed Classification Analysis}
\subsection{Per-Finding Classification Performance}
\label{appendix:class_performance}

Tables \ref{tab:apx_class_wise_performance_openi}-\ref{tab:apx_class_wise_performance_nih} 
compare RadZero and CARZero \cite{carzero} 
in terms of per-finding classification AUCs on
the OpenI, PadChest, and ChestXray14 datasets, respectively.
For PadChest, which has a large number of classes, we evaluated on five representative categories commonly used.
The full names for each abbreviation are provided in Table \ref{tab:abbreviation}.

\begin{table*}[h]
\centering
\small                
\resizebox{1.0\textwidth}{!}{
\begin{tabular}{l|c|c|c|c|c|c|c|c|c|c|c|c|c|c|c|c|c|c|c}
\Xhline{1pt}
Method  & Mean & ATE & CARD & EFF & INFL & MASS & NOD & PNA & PTX & EDE & EMPH & FIB & PLTH & HERN & FX & OPAC & LES & CG & LG \\
\hline\hline
CarZero\cite{carzero} & 0.838 & \textbf{0.859} & 0.933 & \textbf{0.938} & \textbf{0.776} & 0.887 & 0.612 & 0.877 & 0.921 & 0.900 & 0.899 & \textbf{0.917} & 0.822 & 0.953 & \textbf{0.726} & \textbf{0.784} & \textbf{0.976} & 0.658 & 0.621 \\
\textbf{RadZero(224px)} & \textbf{0.851} & 0.850 & \textbf{0.939} & 0.937 & 0.774 & \textbf{0.891} & \textbf{0.653} & \textbf{0.881} & \textbf{0.952} & \textbf{0.910} & 0.925 & 0.900 & \textbf{0.837} & \textbf{0.989} & 0.720 & \textbf{0.784} & 0.970 & 0.700 & 0.700 \\
\textbf{RadZero} & 0.847 & 0.857 & 0.933 & 0.933 & 0.775 & 0.886 & 0.630 & 0.870 & 0.949 & 0.902 & \textbf{0.926} & 0.904 & 0.820 & 0.982 & 0.701 & 0.781 & 0.929 & \textbf{0.731} & \textbf{0.734} \\
\Xhline{1pt}
\end{tabular}}
\caption{Class-wise disease classification results on the OpenI dataset.
}
\label{tab:apx_class_wise_performance_openi}
\end{table*}
\begin{table*}[h]
\centering
\small                
\resizebox{0.7\textwidth}{!}{
\begin{tabular}{l|c|c|c|c|c|c}
\Xhline{1pt}
Method  & Mean & ATE & CARD & CONS & EDE & PNA \\
\hline\hline
CarZero\cite{carzero} & 0.810 & 0.835 & 0.906 & \textbf{0.903} & 0.971 & 0.841 \\
\textbf{RadZero(224px)} & \textbf{0.841} & \textbf{0.839} & 0.917 & 0.902 & \textbf{0.973} & \textbf{0.846} \\
\textbf{RadZero} & \textbf{0.841} & \textbf{0.839} & \textbf{0.920} & 0.899 & 0.972 & 0.831 \\
\Xhline{1pt}
\end{tabular}}
\caption{Class-wise disease classification results on the PadChest dataset.
}
\label{tab:apx_class_wise_performance_padchest}
\end{table*}
\begin{table*}[h!]
\centering
\small                
\resizebox{1.0\textwidth}{!}{
\begin{tabular}{l|c|c|c|c|c|c|c|c|c|c|c|c|c|c|c}
\Xhline{1pt}
Method  & Mean & ATE & CARD & EFF & INFL & MASS & NOD & PNA & PTX & CONS & EDE & EMPH & FIB & PLTH & HERN \\
\hline\hline
CarZero\cite{carzero} & \textbf{0.811} & \textbf{0.819} & 0.852 & \textbf{0.873} & 0.670 & 0.854 & 0.718 & 0.737 & 0.871 & \textbf{0.786} & \textbf{0.884} & \textbf{0.808} & 0.788 & 0.770 & 0.928 \\
\textbf{RadZero(224px)} & 0.807 & 0.796 & \textbf{0.864} & 0.857 & \textbf{0.672} & \textbf{0.859} & \textbf{0.742} & \textbf{0.772} & 0.873 & 0.784 & 0.883 & 0.631 & \textbf{0.807} & \textbf{0.789} & \textbf{0.963} \\
\textbf{RadZero} & 0.804 & 0.792 & 0.863 & 0.854 & 0.669 & 0.837 & 0.727 & 0.766 & \textbf{0.875} & 0.784 & 0.881 & 0.655 & 0.806 & 0.781 & \textbf{0.963} \\
\Xhline{1pt}
\end{tabular}}
\caption{Class-wise disease classification results on the NIH ChestXray14 dataset.
}
\label{tab:apx_class_wise_performance_nih}
\end{table*}

Consistent with the average performance, RadZero generally outperforms or matches CARZero on OpenI and PadChest. 
On ChestXray14, our model performs comparably or better on most pathologies, but a notable drop on emphysema (EMPH) accounts for the overall underperformance on the dataset. 
However, RadZero outperforms CARZero on EMPH in OpenI, indicating that the drop is dataset-driven rather than due to lesion-specific modeling issues. 
These results reaffirm that while RadZero demonstrates strong zero-shot classification, further improvements in generalization remain an important direction.

\subsection{Comparison with Supervised Baselines}

Table \ref{tab:apx_supervised_comparison} compares our zero-shot classification performance against supervised baselines reported in prior work on the CheXpert and ChestXray14 datasets.
For the baselines, the percentages in parentheses indicate the proportion of training data used for supervised training.
Since RadZero is trained solely on MIMIC-CXR, its external test results on CheXpert and ChestXray14 are reported as single zero-shot scores rather than separate results for 1\%, 10\%, and 100\% of the training data.
Note that the results for CheXpert differ from those presented in Table \ref{tab:zero-shot-cls}, as a different test set was used.
Specifically, we followed the test split adopted in \cite{mrm} 
to ensure a fair comparison with other models.
Although both RadZero and RadZero(224px) underperform some fully supervised models trained with 10\% or 100\% of labeled data, they consistently outperform models trained with 1\% supervision.
These results demonstrate that RadZero achieves competitive zero-shot classification performance, consistent with the segmentation results in Table \ref{tab:zero-shot-seg}.

\begin{table*}[h]
\centering
\small
\resizebox{0.9\textwidth}{!}{
\begin{tabular}{l|c|c|c|c|c|c}
\Xhline{1pt}
\multirow{2}{*}{Method} 
& \multicolumn{3}{c|}{CheXpert}
& \multicolumn{3}{c}{ChestXray14} \\
\cline{2-7}
& (1\%) & (10\%) & (100\%) & (1\%) & (10\%) & (100\%) \\
\hline\hline
ConVIRT~\cite{convirt} & 0.870 & 0.881 & 0.881 & - & - & - \\
REFERS~\cite{refers}$^*$ & 0.872 & 0.881 & 0.882 & - & - & - \\
M3AE~\cite{m3ae}$^*$ & 0.862 & 0.873 & 0.879 & - & - & - \\
MGCA~\cite{mgca} & 0.888 & 0.891 & 0.897 & - & - & - \\
MRM~\cite{mrm} & 0.885 & 0.885 & 0.887 & 0.794 & 0.840 & 0.859 \\
\hline
\textbf{RadZero(224px)} 
& \multicolumn{3}{c|}{0.888}
& \multicolumn{3}{c}{0.807} \\
\textbf{RadZero} 
& \multicolumn{3}{c|}{0.889}
& \multicolumn{3}{c}{0.804} \\
\Xhline{1pt}
\end{tabular}}
\caption{Comparison of zero-shot classification performance of RadZero against supervised baselines on CheXpert and ChestXray14 datasets.
Results for models marked with an asterisk ($^*$) are taken from \cite{mrm}, whereas the results for all other models are reported in their respective papers.
The percentages in parentheses indicate the proportion of training data used for supervised training.
}
\label{tab:apx_supervised_comparison}
\end{table*}

\section{Additional Details}

\subsection{Abbreviations}
\label{appendix:abbreviation}

Table \ref{tab:abbreviation} lists the abbreviations used in this paper for lesions and anatomical regions. 
The left column shows the abbreviated terms, and the right column gives their description.

\begin{table*}[h]
\centering
\begin{minipage}[t]{0.46\textwidth}
\scalebox{.8}{
\begin{tabular}{c|>{\centering\arraybackslash}p{5.3cm}}
\Xhline{1pt}
Abbreviation  & Description \\
\hline\hline
ATE & Atelectasis \\
CALC & Calcification \\
CARD & Cardiomegaly \\
CG & Calcified Granuloma \\
CONS & Consolidation \\
EDE & Pulmonary Edema \\
EFF & Effusion \\
EMPH & Emphysema \\
FIB & Fibrosis \\
FX & Fracture \\
HERN & Hernia \\
INFL & Infiltration \\
LES & Lesion \\
LG & Lung Granuloma \\
MASS & Mass \\
NOD & Nodule \\
OPAC & Opacity \\
PLTH & Pleural Thickening \\
PNA & Pneumonia \\
PTX & Pneumothorax \\
\Xhline{1pt}
\end{tabular}}
\caption*{(a) Lesion abbreviations}
\end{minipage}
\hspace{0.05\textwidth} 
\begin{minipage}[t]{0.46\textwidth}
\scalebox{.8}{
\begin{tabular}{c|>{\centering\arraybackslash}p{5.3cm}}
\Xhline{1pt}
Abbreviation  & Description \\
\hline\hline
UL & Upper Lobe \\
ML & Mid Lobe \\
LL & Lower Lobe \\
Rclav & Right Clavicle \\
Lclav & Left Clavicle \\
RULZ & Right Upper Lung Zone \\
RMLZ & Right Mid Lung Zone \\
RLLZ & Right Lower Lung Zone \\
LULZ & Left Upper Lung Zone \\
LMLZ & Left Mid Lung Zone \\
LLLZ & Left Lower Lung Zone \\
RCPA & Right Costophrenic Angle \\
LCPA & Left Costophrenic Angle \\
HD & Hemidiaphragm \\
RHD & Right Hemidiaphragm \\
LHD & Left Hemidiaphragm \\
\Xhline{1pt}
\end{tabular}}
\caption*{(b) Anatomical region abbreviations}
\end{minipage}

\caption{Abbreviations for lesions and anatomical regions.}

\label{tab:abbreviation}
\end{table*}

\subsection{Prompt for finding-sentence extraction.}
\label{appendix:prompt}

\begin{figure*}[h]
\centering
\includegraphics[width=1.0\textwidth]{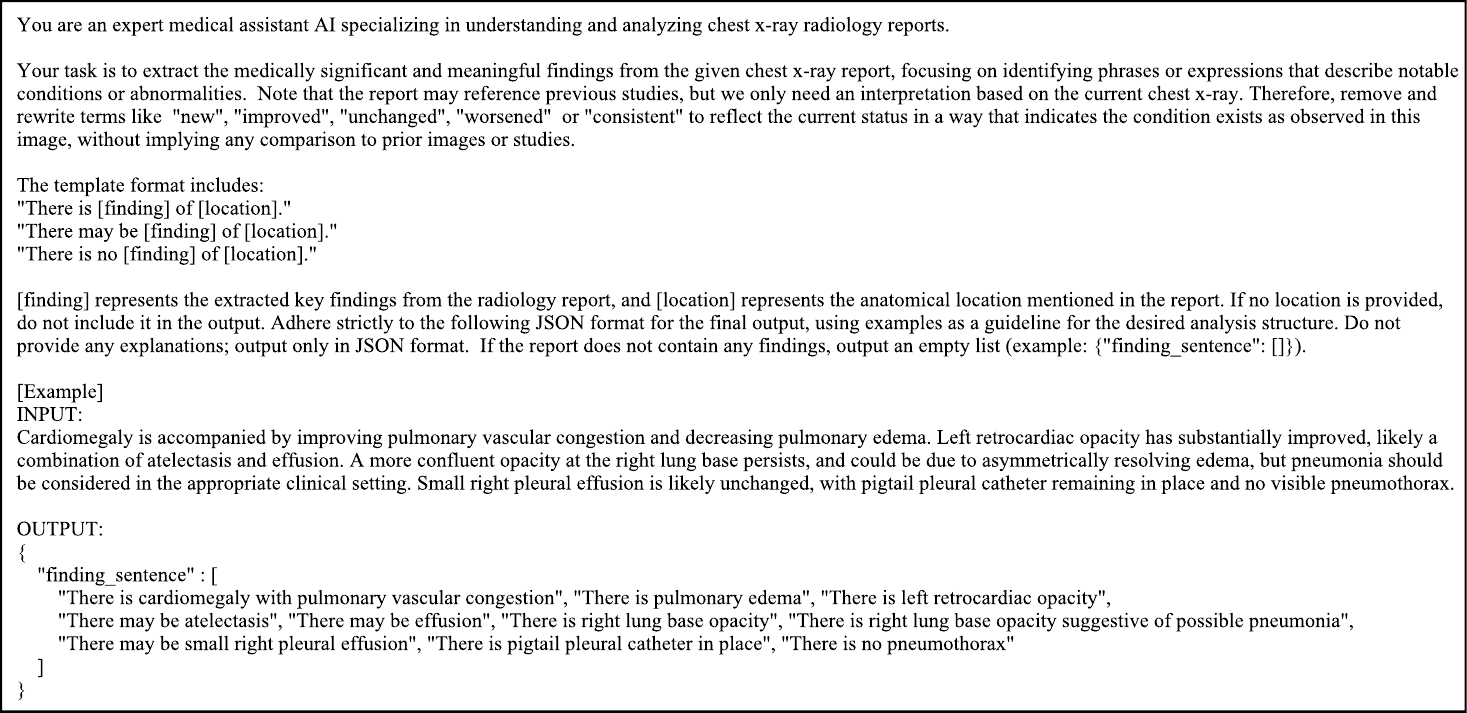}
\caption{Prompt design for extracting finding-sentences with LLM.}
\label{fig:prompt}
\end{figure*}

As shown in Figure \ref{fig:prompt}, the prompt instructs the LLM to extract clinically relevant minimal semantic units in the form of sentences from radiology reports.
Finding-sentences are standardized through prompt alignment to follow a ``There is'' format, with a one-shot example enhancing extraction accuracy and guiding the model to identify both findings and their corresponding anatomical locations in a structured manner.

\section{Linguistic Robustness and Generalization Analysis}

\begin{wraptable}{r}{0.6\textwidth}  
\centering
\scriptsize
\setlength{\tabcolsep}{6pt}
\scalebox{0.85}{
\begin{tabular}{
  >{\raggedright\arraybackslash}p{0.18\textwidth}  
  >{\raggedright\arraybackslash}p{0.35\textwidth}  
  >{\raggedleft\arraybackslash}p{0.07\textwidth}   
}
\toprule
\textbf{Query} & \textbf{Retrieved Sentences (Top-k)} & \textbf{Similarity} \\
\midrule

\multirow{6}{=}{(a) \textbf{the lungs are clear}}%
\retr{The lungs are clear}{1.000}
\retr{Lungs are clear}{0.994}
\retr{The airways are clear}{0.993}
\retr{There is clear lungs}{0.985}
\retr{There are clear lungs}{0.983}
\retr{The chest is clear}{0.982}

\midrule

\multirow{6}{=}{(b) \textbf{pleural effusion in the right lower lung}}%
\retr{There is pleural effusion in the right lower lung}{0.993}
\retr{There is a pleural effusion in the right lower lung}{0.978}
\retr{There is pleural effusion of the right lower lung}{0.976}
\retr{There is effusion in the right lower lung}{0.968}
\retr{There is a pleural effusion of the right lower lung}{0.965}
\retr{There is right lower lung pleural effusion}{0.953}

\midrule

\multirow{6}{=}{(c) \textbf{there is fibrosis}}%
\retr{There is fibrosis}{1.000}
\retr{There is probable fibrosis}{0.937}
\retr{There is lung fibrosis}{0.937}
\retr{There is chronic fibrosis}{0.934}
\retr{There is a component of fibrosis}{0.924}
\retr{There is fibrotic disease}{0.917}

\midrule

\multirow{3}{=}{(d) \textbf{pleural effusion in RLL \textcolor{red}{(unseen)}}}%
\retr{There is pleural effusion with drainage}{0.753}
\retr{There is slight decrease in pleural fluid}{0.725}
\retr{There is collecting pleural fluid}{0.708}

\bottomrule
\end{tabular}}
\caption{
Examples of similarity search results using RadZero’s text encoder.
Cell colors indicate cosine similarity: high (blue) $\geq 0.95$, medium (yellow) $\geq 0.85$, and low (red) $< 0.85$.
}
\label{tab:text_encoder_similarity}
\end{wraptable}

To evaluate the linguistic robustness of RadZero 
to variations in
report templates and phrasing patterns,
we conducted a series of qualitative analyses using similarity search within a vector database of training sentences encoded by RadZero’s text encoder.
In Table \ref{tab:text_encoder_similarity} (a), given a query sentence ``the lungs are clear'', the model retrieved multiple semantically similar sentences with varied phrasing and syntactic structures. 
These results indicate that the encoder captures semantic equivalence beyond fixed textual templates, such as recognizing ``the chest is clear'' or ``the airways are clear'' as close variants.

We further tested syntactic diversity using queries with complex structures (e.g., ``pleural effusion in the right lower lung''). As shown in Table \ref{tab:text_encoder_similarity} (b), the retrieved results included expressions such as ``right lower lung pleural effusion'' and ``pleural effusion of the right lower lung,'' demonstrating the model’s robustness to grammatical variations.
Similarly, queries evaluating lexical flexibility (e.g., ``there is fibrosis'' in Table \ref{tab:text_encoder_similarity} (c)) showed that RadZero identifies semantically related expressions such as ``chronic fibrosis'' and ``fibrotic disease,'' suggesting strong generalization across vocabulary variations.
These findings collectively suggest that RadZero is not confined to specific textual templates or vocabularies and generalizes well across diverse linguistic patterns, thereby mitigating the risk of overfitting to particular report styles.

However, we observed that embeddings may be less accurate for report expressions that were rare or absent in the training data. 
One notable case involves abbreviations-for example, ``RLL'' (right lower lung)-which occurred infrequently in the original reports. 
As shown in Table \ref{tab:text_encoder_similarity} (d), such unseen abbreviations tend to yield weaker semantic alignment.
To address this limitation, non-standardized or abbreviated expressions can be rewritten into complete finding-sentences using LLM-based normalization, which helps maintain robustness across diverse report formats.

\section{Statistical Significance of Main Results}
\label{appendix:statistical_significance}

To assess run-to-run variability and verify the statistical reliability of the reported performance, we conducted additional experiments across five random seeds for all tasks. 
For each model and task, we report the mean and standard deviation of the evaluation metrics. 

\begin{table*}[h]
  \centering
  \small                
  \resizebox{\textwidth}{!}{
    \begin{tabular}{l c c c c c c c c c c c c c}
      \toprule
      \multirow{2}{*}{Method} 
            & \multicolumn{8}{c}{Classification} 
            & \multicolumn{2}{c}{Grounding} 
            & \multicolumn{2}{c}{Segmentation} \\
      \cmidrule(lr){2-9} \cmidrule(lr){10-11} \cmidrule(lr){12-13}
            & OI & PC & PC20 & CXR14 & CXP & CXD10 & SIIM & RSNA 
            & CXD10 & MS-CXR & SIIM & RSNA \\
      \midrule
      \textbf{RadZero} 
            & 0.847 & 0.841 & 0.871 & 0.804 & 0.900 & 0.787 & 0.924 & 0.834 
            & 0.622 & 0.844 & 0.171 & 0.546\\
   
      \midrule

      Mean~($\pm$) 
            & 0.848~($\pm$) & 0.840~($\pm$)  & 0.869~($\pm$)  & 0.802~($\pm$)  & 0.899~($\pm$)  & 0.787~($\pm$) & 0.918~($\pm$)  & 0.843~($\pm$) 
            & 0.601~($\pm$)  & 0.845~($\pm$)  & 0.164~($\pm$)  & 0.549~($\pm$)  \\
      std~ 
            & 0.0016 & 0.0011 & 0.0040 & 0.0016 & 0.0019 & 0.0043 & 0.0063 & 0.0072 
            & 0.0130 & 0.0143 & 0.0080 & 0.0023 \\
      
      \bottomrule
    \end{tabular}
  }
  \caption{Mean and standard deviation of main results over five runs.}
  \label{tab:apx_statistical_significance}
\end{table*}

As shown in Table \ref{tab:apx_statistical_significance}, 
the variance across runs is relatively small, indicating that the observed improvements are consistent and not due to random fluctuations.
These results confirm the stability and robustness of RadZero across different random initializations.

\begin{wrapfigure}{r}{0.5\textwidth}
\vspace{-25pt}  
    \centering
    \includegraphics[width=\linewidth]{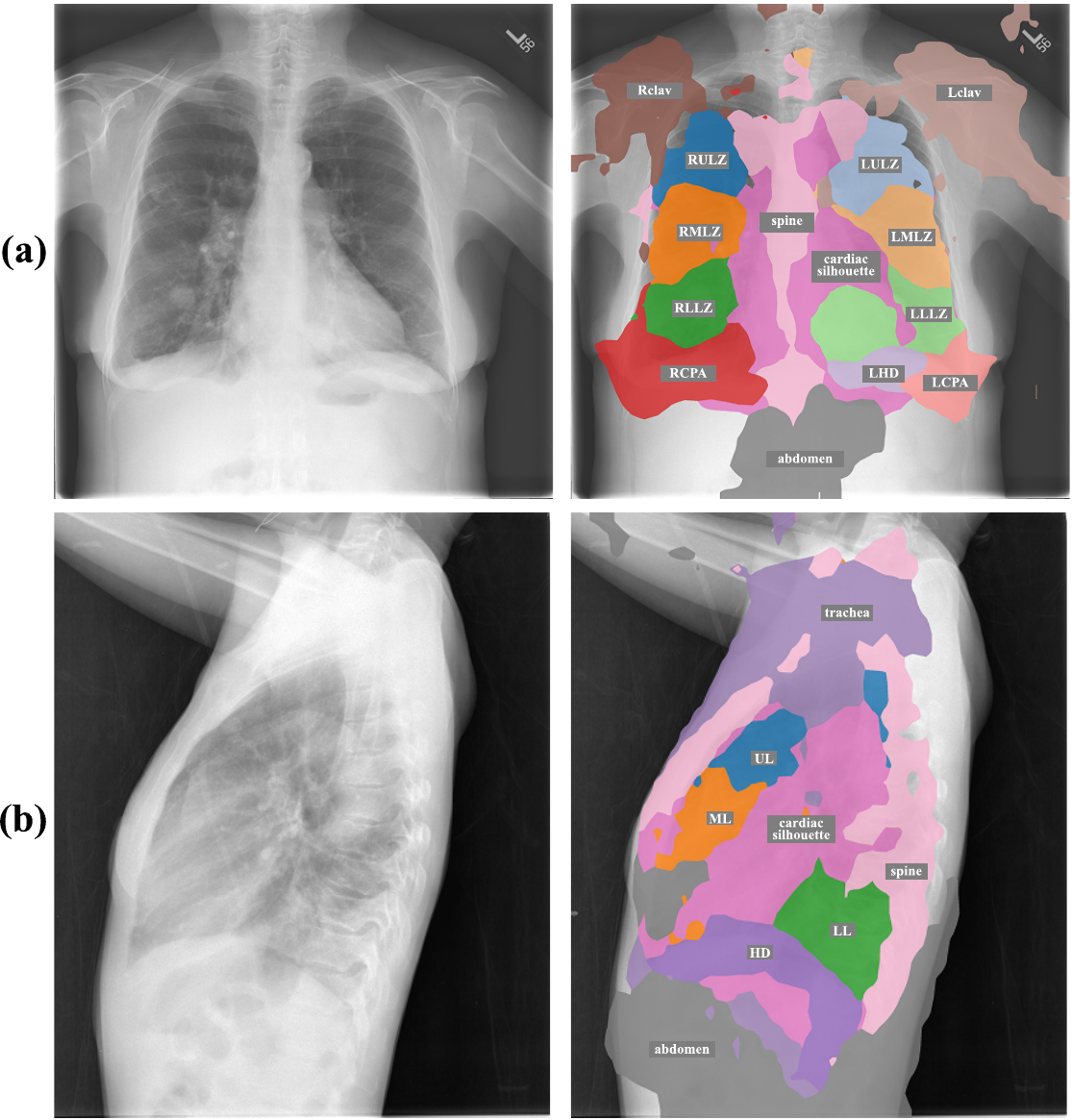}
    \caption{
        Open-vocabulary semantic segmentation for anatomical regions.
        The CXR images are sourced from Open-I. The segmentation threshold was set to 0.4.
    }
    \label{fig:open vocab segmentation apx}
\end{wrapfigure}

\section{Additional Visualization Results}
\label{sec:additional_visualization}

\begin{figure*}[htbp!]
\centering

\begin{minipage}[t]{0.95\textwidth}
    \centering
    \includegraphics[width=0.9\linewidth]{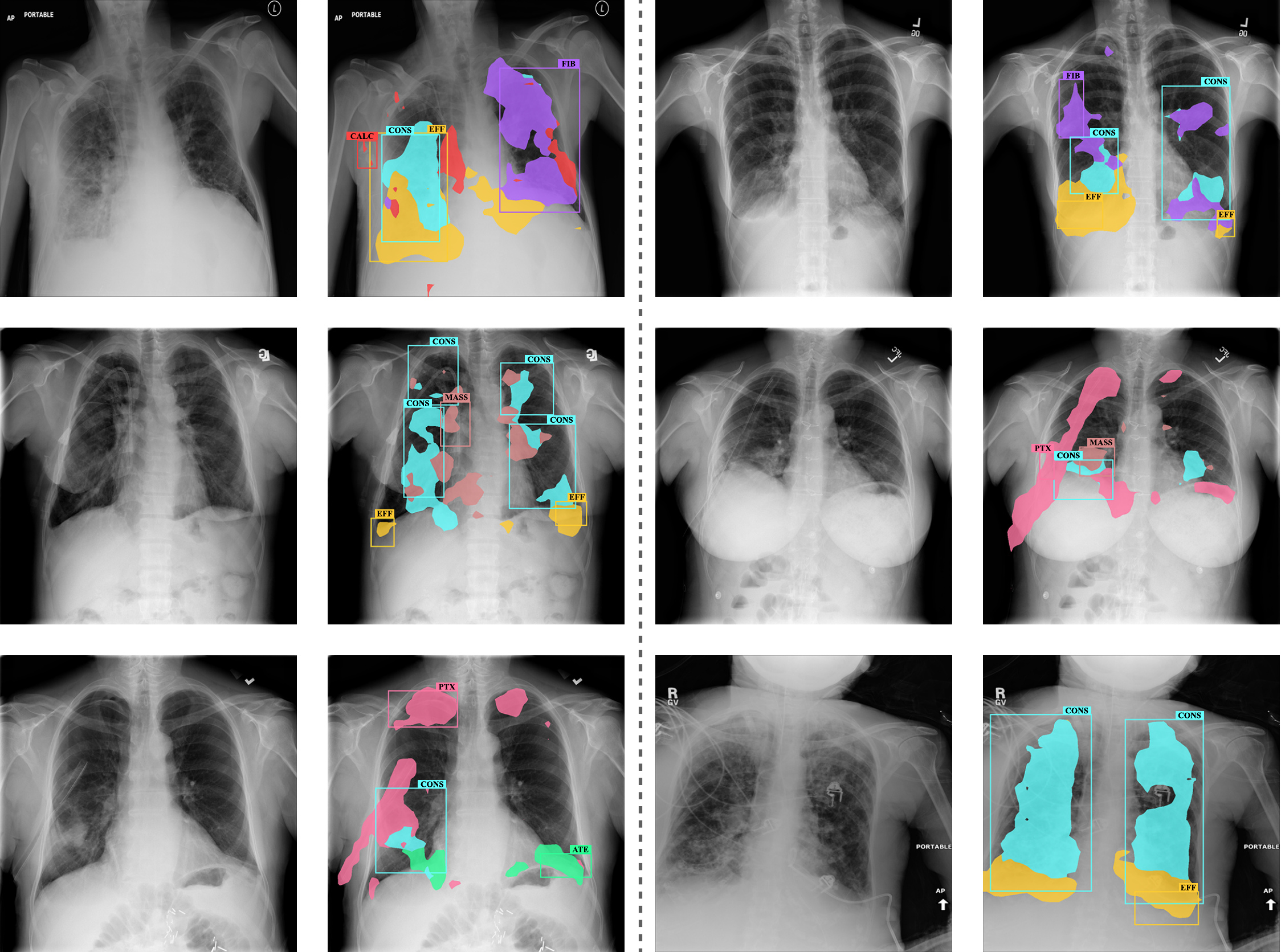}
    \caption{
    Open-vocabulary semantic segmentation for findings.
    The CXR images and bounding box labels are sourced from ChestXDet10. 
    The segmentation threshold was set to 0.7.
    }
    \label{fig:findings segmentation apx}
\end{minipage}

\vspace{1.5em}  

\begin{minipage}[t]{0.95\textwidth}
    \centering
    \includegraphics[width=0.9\linewidth]{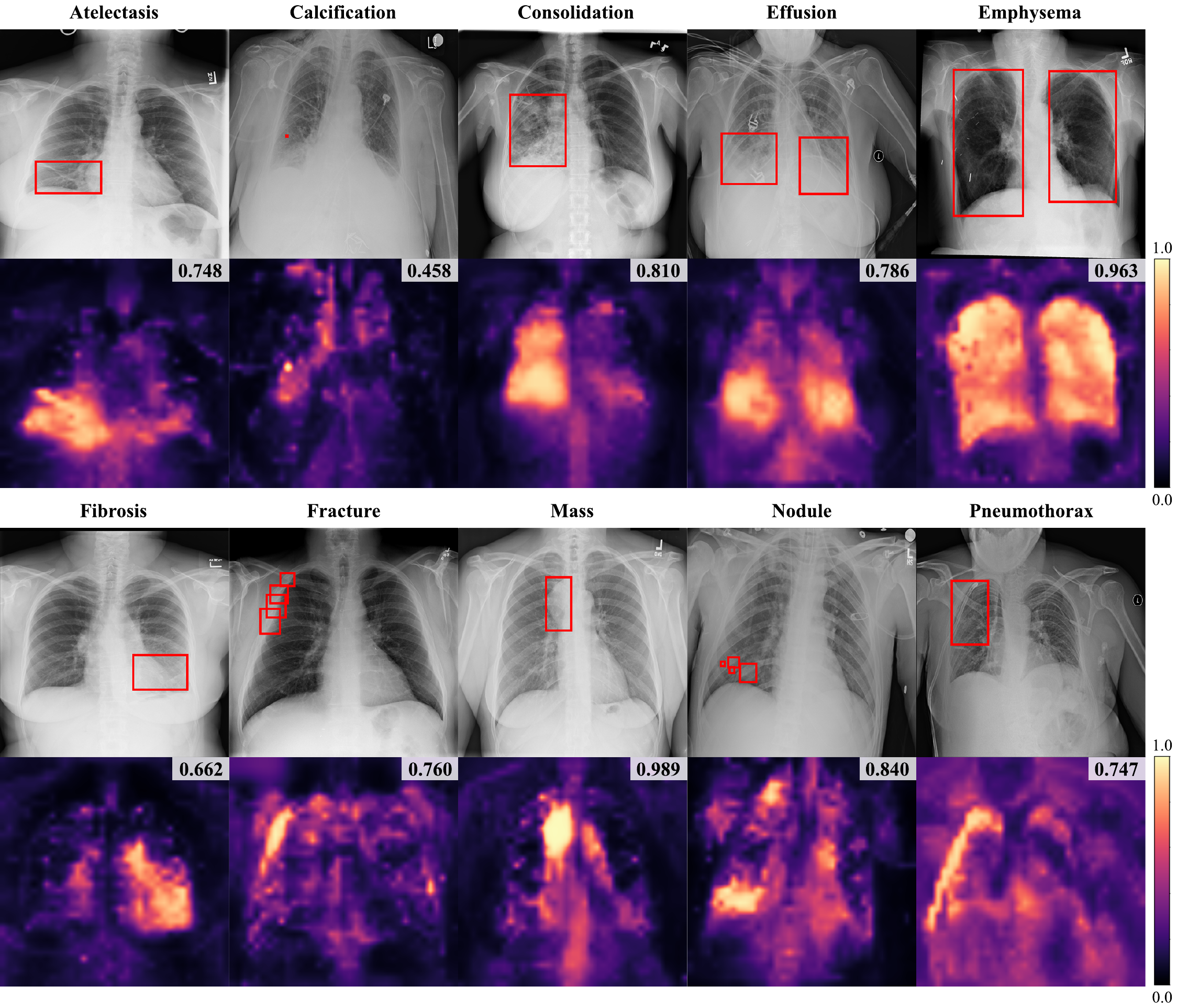}
    \caption{
    Visualization of VL similarity maps generated by RadZero for 10 findings on the ChestXDet10 dataset.
    Red boxes indicate ground truth bounding boxes. 
    The similarity probability \( \hat{l} \) is shown in the top-right corner of each map.
    }
    \label{fig:findings similarity map}
\end{minipage}

\end{figure*}

Figure \ref{fig:open vocab segmentation apx} depicts segmentation of anatomical regions, which, while not perfect, generally align with appropriate locations.
Figure \ref{fig:findings segmentation apx} presents examples demonstrating RadZero’s potential for open-vocabulary semantic segmentation, including additional lesion types such as mass, fibrosis, and calcification.
Full names of lesion and anatomical region abbreviations are provided in Table \ref{tab:abbreviation}.

Figure \ref{fig:findings similarity map} presents VL similarity maps for 10 different findings of the ChestXDet10 dataset, following the pipeline in Sec. \ref{sec:sim-map}. 
The brightest regions in the map
align well with the bounding boxes, even for multiple or small lesions.  
The similarity probability was above 0.5 for all findings except calcification. While the model correctly localized calcifications, the activated regions appeared as small bright spots, leading to a lower similarity probability of 0.45 due to the weighted sum calculation. 
This highlights a limitation of RadZero, suggesting the need for further refinement in future work.

\clearpage

\section*{NeurIPS Paper Checklist}

\begin{enumerate}

\item {\bf Claims}
    \item[] Question: Do the main claims made in the abstract and introduction accurately reflect the paper's contributions and scope?
    \item[] Answer: \answerYes{} 
    \item[] Justification: 
    The abstract and introduction accurately reflect the paper’s scope by highlighting RadZero and its core component, VL-CABS, which computes text–image similarity for interpretable, fine-grained vision-language alignment. They also clearly describe RadZero’s zero-shot capability across multiple tasks.

\item {\bf Limitations}
    \item[] Question: Does the paper discuss the limitations of the work performed by the authors?
    \item[] Answer: \answerYes{} 
    \item[] Justification: The paper explicitly discusses RadZero’s limitations in each section including Sec. \ref{sec:conclusion} Conclusion:
    performance gaps on certain datasets (Sec. \ref{sec:zero-shot evaluation}),
    prediction of regions extending beyond the ground truth bounding box in open-vocabulary segmentation (Sec. \ref{sec:ov-seg}),
    and reliance on a pre-trained vision encoder (Sec. \ref{sec:ablation}), 
    suggesting directions for future improvement.

\item {\bf Theory assumptions and proofs}
    \item[] Question: For each theoretical result, does the paper provide the full set of assumptions and a complete (and correct) proof?
    \item[] Answer: \answerNA{} 
    \item[] Justification: This work mainly includes empirical contributions.

    \item {\bf Experimental result reproducibility}
    \item[] Question: Does the paper fully disclose all the information needed to reproduce the main experimental results of the paper to the extent that it affects the main claims and/or conclusions of the paper (regardless of whether the code and data are provided or not)?
    \item[] Answer: \answerYes{} 
    \item[] Justification: We provide detailed experimental configurations in Sections 
    \ref{sec:impl_detail} and Appendices
    \ref{sec:training}, \ref{appendix:prompt}.

\item {\bf Open access to data and code}
    \item[] Question: Does the paper provide open access to the data and code, with sufficient instructions to faithfully reproduce the main experimental results, as described in supplemental material?
    \item[] Answer: \answerYes{} 
    \item[] Justification:
    Our experiments are all conducted on publicly accessible datasets, 
    and the details about the datasets used are described in Sections 
    \ref{sec:training-data} (training data) and
    \ref{subsec:eval_dataset} (test data).
    For experiment implementation, we follow the official code of exisiting works, all code can be found in their official GitHub repository.

\item {\bf Experimental setting/details}
    \item[] Question: Does the paper specify all the training and test details (e.g., data splits, hyperparameters, how they were chosen, type of optimizer, etc.) necessary to understand the results?
    \item[] Answer: \answerYes{} 
    \item[] Justification:  We provide detailed for 
    experimental configurations in Section \ref{sec:impl_detail} and Appendices \ref{sec:training}, \ref{appendix:prompt},
    and for the data split in Sections 
    \ref{sec:training-data} (training data) and
    \ref{subsec:eval_dataset} (test data).

\item {\bf Experiment statistical significance}
    \item[] Question: Does the paper report error bars suitably and correctly defined or other appropriate information about the statistical significance of the experiments?
    \item[] Answer: \answerYes{} 
    \item[] Justification: We report mean and standard deviation across five random seeds in Appendix \ref{appendix:statistical_significance}, ensuring statistical significance and reliability of the results.

\item {\bf Experiments compute resources}
    \item[] Question: For each experiment, does the paper provide sufficient information on the computer resources (type of compute workers, memory, time of execution) needed to reproduce the experiments?
    \item[] Answer: \answerYes{} 
    \item[] Justification: The information on the computation resources is described in Appendix \ref{sec:training}.
    
\item {\bf Code of ethics}
    \item[] Question: Does the research conducted in the paper conform, in every respect, with the NeurIPS Code of Ethics \url{https://neurips.cc/public/EthicsGuidelines}?
    \item[] Answer: \answerYes{} 
    \item[] Justification: This research was conducted in the paper conform, in every respect, with the NeurIPS Code of Ethics.

\item {\bf Broader impacts}
    \item[] Question: Does the paper discuss both potential positive societal impacts and negative societal impacts of the work performed?
    \item[] Answer: \answerYes{} 
    \item[] Justification: The potential societal impacts are mentioned in Sections \ref{sec:intro} and \ref{sec:conclusion}.
    
\item {\bf Safeguards}
    \item[] Question: Does the paper describe safeguards that have been put in place for responsible release of data or models that have a high risk for misuse (e.g., pretrained language models, image generators, or scraped datasets)?
    \item[] Answer: \answerNA{} 
    \item[] Justification: This work does not release data or models that have a high risk for misuse.

\item {\bf Licenses for existing assets}
    \item[] Question: Are the creators or original owners of assets (e.g., code, data, models), used in the paper, properly credited and are the license and terms of use explicitly mentioned and properly respected?
    \item[] Answer: \answerYes{} 
    \item[] Justification: Please refer to Sections \ref{sec:training-data} and \ref{subsec:eval_dataset}.

\item {\bf New assets}
    \item[] Question: Are new assets introduced in the paper well documented and is the documentation provided alongside the assets?
    \item[] Answer: \answerNA{} 
    \item[] Justification: There is no new assets released in this work.

\item {\bf Crowdsourcing and research with human subjects}
    \item[] Question: For crowdsourcing experiments and research with human subjects, does the paper include the full text of instructions given to participants and screenshots, if applicable, as well as details about compensation (if any)? 
    \item[] Answer: \answerNA{} 
    \item[] Justification: This work has no human subjects.

\item {\bf Institutional review board (IRB) approvals or equivalent for research with human subjects}
    \item[] Question: Does the paper describe potential risks incurred by study participants, whether such risks were disclosed to the subjects, and whether Institutional Review Board (IRB) approvals (or an equivalent approval/review based on the requirements of your country or institution) were obtained?
    \item[] Answer: \answerNA{} 
    \item[] Justification: This work has no human subjects.

\item {\bf Declaration of LLM usage}
    \item[] Question: Does the paper describe the usage of LLMs if it is an important, original, or non-standard component of the core methods in this research? Note that if the LLM is used only for writing, editing, or formatting purposes and does not impact the core methodology, scientific rigorousness, or originality of the research, declaration is not required.
    \item[] Answer: \answerYes{} 
    \item[] Justification: 
    This work have used a publicly available LLM for data processing.
    To prevent any potential misuse, the LLM was executed on a secure, private server.

\end{enumerate}

\end{document}